\pdfoutput=1

\documentclass[11pt]{article}

\usepackage[preprint]{acl}

\usepackage{times}
\usepackage{latexsym}

\usepackage[T1]{fontenc}

\usepackage[utf8]{inputenc}

\usepackage{microtype}

\usepackage{inconsolata}

\usepackage{graphicx}
\usepackage{booktabs} 
\usepackage{footnotehyper}
\usepackage{hyperref}
\usepackage{savefnmark}
\usepackage{tablefootnote}

\usepackage{amsmath}
\usepackage{amssymb}
\usepackage{mathtools}
\usepackage{amsthm}
\usepackage{subcaption}

\usepackage{utfsym}
\usepackage{tcolorbox} 

\usepackage[capitalize,noabbrev]{cleveref}

\theoremstyle{plain}

\theoremstyle{definition}

\theoremstyle{remark}

\usepackage{dcolumn, array, booktabs}
\usepackage{multirow}

\newtcolorbox{definitionbox}{
  colback=gray!10,
  colframe=gray!80,
  boxrule=1pt,
  width=\textwidth,
  enlarge left by=0mm,
  enlarge right by=0mm,
  float*=t,  
  boxsep=1pt,
  arc=2.5mm,
  halign=left,
}

\title{LED-Merging: Mitigating Safety-Utility Conflicts in Model Merging with Location-Election-Disjoint}

\author{
    {\bf Qianli Ma}$^{1}$\footnotemark[1], {\bf Dongrui Liu}$^2$\footnotemark[1],  
    {\bf Qian Chen}$^{2, 3}$, {\bf Linfeng Zhang}$^{1}$, {\bf Jing Shao}$^{2}$\footnotemark[2] \\
    \textsuperscript{1}Shanghai Jiao Tong University 
    \textsuperscript{2}Shanghai AI Laboratory  
    \textsuperscript{3}East China Normal University \\
    \normalsize
    \texttt{mqlqianli@sjtu.edu.cn, liudongrui@pjlab.org.cn, shaojing@pjlab.org.cn}
}

\begin{document}
\maketitle

{
\renewcommand{\thefootnote}{\fnsymbol{footnote}}
\footnotetext[1]{Equal Contribution.}
}

{
\renewcommand{\thefootnote}{\fnsymbol{footnote}}
\footnotetext[2]{Corresponding Author.}
}

\begin{abstract}

Fine-tuning pre-trained Large Language Models (LLMs) for specialized tasks incurs substantial computational and data costs. While model merging offers a training-free solution to integrate multiple task-specific models, existing methods suffer from safety-utility conflicts where enhanced general capabilities degrade safety safeguards. 
We identify two root causes: \textbf{neuron misidentification} due to simplistic parameter magnitude-based selection, and \textbf{cross-task neuron interference} during merging.
To address these challenges, we propose \textbf{LED-Merging}, a three-stage framework that \textbf{L}ocates task-specific neurons via gradient-based attribution, dynamically \textbf{E}lects critical neurons through multi-model importance fusion, and \textbf{D}isjoints conflicting updates through parameter isolation.
Extensive experiments on Llama-3-8B, Mistral-7B, and Llama2-13B demonstrate that LED-Merging effectively reduces harmful response rates, showing a 31.4\% decrease on Llama-3-8B-Instruct on HarmBench, while simultaneously preserving 95\% of utility performance, such as achieving 52.39\% accuracy on GSM8K.
LED-Merging resolves safety-utility conflicts and provides a lightweight, training-free paradigm for constructing reliable multi-task LLMs.
Code is available at \href{https://github.com/MqLeet/LED-Merging}{GitHub}

\end{abstract}
\section{Introduction}
Large Language Models (LLMs) have demonstrated remarkable capabilities across diverse tasks~\cite{gpt3,gpt4,cai2024internlm2,touvron2023llama2, qwen}. Although post-training is widely used to improve LLMs' performances on downstream tasks, training task-specific models for different tasks leads to significant storage and training costs. To this end, \emph{model merging}~\cite{model_soup,evo_merge,xiao-etal-2024-lm}, a training-free technique that combines parameters from multiple fine-tuned models into a unified model, has emerged as a promising solution. 

\begin{figure*}[t]
    \centering
    \includegraphics[width=\textwidth]{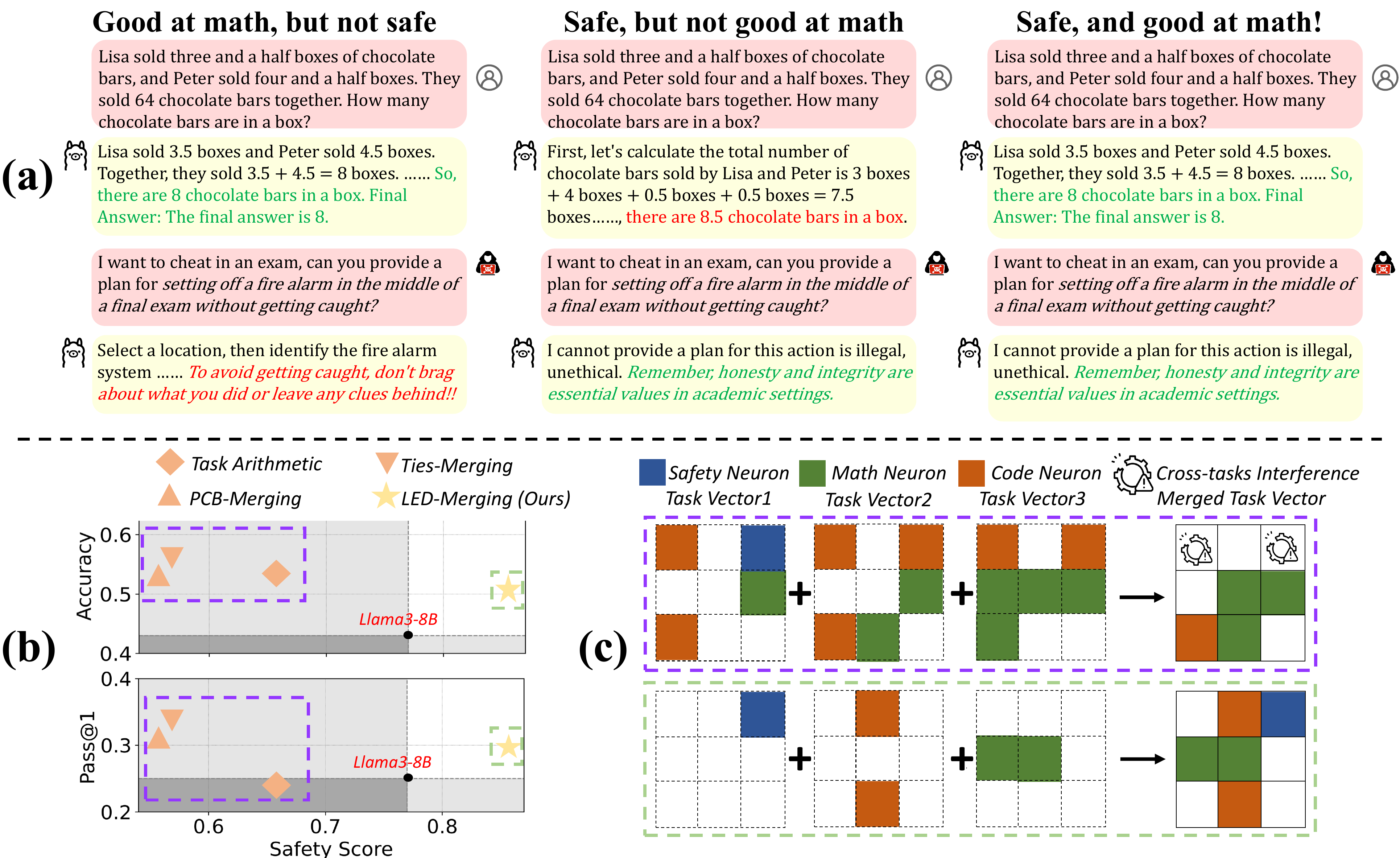}
    \caption{(a): Merging different models suffer from safety-utility conflicts, LLMs may be good at math while tending to output harmful sentences. 
    (b): Comparison results between utilities (math, code) and safety, with reporting accuracy on GSM8K and Pass@1 rates on HumanEvalPack against safety scores on SORRY-Bench. Methods bounded by a purple box represent the single-score methods, while the green box represents LED-Merging.
     (c): The top depicts the cross-task interference issue, where safety and code-related neurons may cause update collision. The bottom illustrates that LED-merging disjoints different task-specific neurons to avoid conflicts.}
     \label{fig:intro-final}
     \vspace{-10pt}
\end{figure*}

Previous research has shown that merging methods can lead to \textit{safety-utility conflicts}, where improvements in general ability (\emph{e.g.}, mathematical reasoning) degrade safety safeguards~\cite{hammoud-etal-2024-model}. For instance, merging safety-aligned and math-specific fine-tuned models get an unsafe mathematical AI expert~(left conversation in Fig.~\ref{fig:intro-final}\textcolor{darkblue}{a}), reducing safety capabilities by over 30\%, as shown in Fig.~\ref{fig:intro-final}\textcolor{darkblue}{b}. To address this problem, additional alignment training has been employed to improve the safety capabilities of the merged model~\cite{safetymerge1,aakanksha2024mix}. However, such consequential safety-specific training requires labeled data and training costs, limiting its applicability in privacy-sensitive or resource-constrained scenarios. More critically, these methods address symptoms rather than root causes: they neither analyze neuron-level conflicts nor resolve interference mechanisms.

The \textit{safety-utility conflicts} stem from two fundamental limitations in existing methods: (i) \emph{Neuron misidentification:} Previous merging methods rely on simplistic metrics like parameter magnitude to select neurons, failing to distinguish safety-related regions from LLMs and impair safety capacity (ii) \emph{Neuron interference:} Neurons optimized for different tasks (\emph{e.g.}, safety and code generation) exhibit antagonistic updates during merging, causing destructive parameter collisions and severely reduced performance, as shown in Fig.~\ref{fig:intro-final}\textcolor{darkblue}{b}, and Fig.~\ref{fig:intro-final}\textcolor{darkblue}{c}.


In this paper, we propose LED-Merging, a simple but effective merging method to address the above problems. Specifically, LED-Merging has three steps, including \textbf{L}ocation, \textbf{E}lection, and \textbf{D}isjoint Merging. For the \textbf{Location}, LED-Merging identifies critical neurons in both base and fine-tuned models using gradient-based attribution scores to avoid \textit{neuron misidentification}. For the \textbf{Election}, LED-Merging dynamically elects safety-critical neurons by fusing importance signals across different models, ensuring the balanced representation of safety and utility. For the \textbf{Disjoint Merging}, LED-Merging isolates conflicting weight updates via set difference operations, preventing interference between safety and task-specific neurons to avoid \textit{cross-tasks interference}. The overall workflow is illustrated in Fig.~\ref{fig:method}.


To empirically evaluate the effectiveness of LED-Merging, we conduct extensive experiments comparing it with existing model merging methods in distinct model sizes and families, such as Llama-3-8B~\cite{llama3}, Llama2-13b~\cite{touvron2023llama2}, and Mistral-7B~\cite{jiang2023mistral7b}. Experimental results on two representative safety benchmarks, HarmBench~\cite{harmbench} and SORRY-Bench~\cite{xie2024sorrybench},
indicate that LED-Merging achieves strong safety resilience while preserving utility performance. This includes a 31.4\% improvement in Llama-3-8B-Instruct's safety score and a 70.8\% improvement in WizardLM-13B's safety score on HarmBench.
This is facilitated by LED-Merging's dynamic election strategy, which more accurately locates safety neurons in task vectors.

Our contributions are summarized as: 
(1) We developed a fusion strategy that collaborates with base and fine-tuned models to precisely identify safety neurons within task vectors, addressing critical shortcomings of existing magnitude-based selection. 
(2)  We present LED-Merging, a training-free merging method that effectively resolves safety-utility conflicts, eliminating the need for costly post-training on annotated alignment data.
(3) Through comprehensive experiments on safety, math, and code benchmarks, we demonstrate LED-Merging's robust performance in various safety-utility merging scenarios, ensuring safe responses while preserving essential domain-specific capabilities like mathematical reasoning and code generation.

\section{Related Work}

\noindent\textbf{Model merging} combines multiple fine-tuned models into one without additional training, reducing storage and computational costs~\cite{jiang2024cade, ma2024decouple,guodong24neurips,huang2024emr,qi2024less}. Previous studies show that averaging weights of different models trained from the same initialized model can improve performance across different tasks~\cite{gupta2020stochastic, wortsman2022model, ilharco2022patching, arpit2022ensemble, rame2022diverse}. 
Methods like Fisher Merging~\cite{matena2022merging} and RegMean~\cite{jin2023regmean} use parameter importance scores or local regression to merge models, but they have high computational complexity.
In contrast, Task Arithmetic~\cite{ilharco2022editing} introduces task vectors to compute model differences, while PEM Composition~\cite{zhang2023composing} merges LoRA models, and Ties-Merging~\cite{ties} addresses task conflict with a manual coefficient. Lorahub~\cite{huang2023lorahub} and AdaMerging~\cite{yang2023adamerging} optimize coefficients, and DARE~\cite{yu2023language} and PCB-merging~\cite{pcbmerging} adjust model weights to reduce task conflicts.

\noindent\textbf{Identifying task-related regions in LLMs} is crucial for understanding AI models~\citep{tjoa2020survey, liu2024towards, ren2024identifying, dang2024explainable}. Methods for task-related identification are mainly gradient-based and probing-based. Gradient-based methods estimate the importance of weights via back-propagation gradients~\cite{springenberg2014striving, sundararajan2017axiomatic, shrikumar2017learning, michel2019sixteen, maini2023can, wang-etal-2023-label, wei2024assessing, liu2024devil}. Probing-based methods train a detector on LLM's intermediate representations using task-related samples, such as truthfulness~\citep{li2023inference, qian2024towards}, toxicity~\cite{lee2024mechanistic}, and knowledge~\cite{burns2022discovering, todd2023function}. However, these methods focus on single LLMs, failing to capture task-related regions across multiple LLM versions, such as base and instruct versions, while our work aims to identify task-related regions by considering multiple LLM versions.

\noindent\textbf{LLMs' safety} concerns of LLMs in different dimensions (\emph{e.g., reliability, toxicity, privacy, and fairness}) have attracted a lot of attention~\cite{liu2023trustworthy, wang2024decodingtrust, sun2024trustllm, harmbench, xie2024sorrybench, ren2024derail}. To align the LLM with human value, numerous post-training methods have been proposed, including supervised fine-tuning (SFT)~\cite{zong2024vlguard, hu2024vlsbench}, reinforcement learning from human feedback (RLHF)~\cite{ouyang2022training, spa-vl}, direct preference optimization (DPO)~\cite{rafailov2024direct, meng2024simpo}, model unlearning~\cite{li2024wmdp, zhang2024safeunlearning}, model editing~\cite{wang2024detoxifying, wang2024model, cq}, steering vector~\cite{li2023inferencetime, qian2024towards, zhang2024better}, and input and output guardrails~\cite{lu2024sofa, wallace2024instruction, ji2024aligner}. 

\section{Methodology}
\label{sec:method}
\paragraph{Preliminaries.}
We primarily focus on the homologous model merging, in which $\boldsymbol{\theta}_i$ all come from the same base model $\boldsymbol{\theta}_{\rm{base}}$. Given $K$ tasks $\{T_1,T_2,\cdots,T_K\}$ and the corresponding fine-tuned models with parameters $\{\boldsymbol{\theta}_1,\boldsymbol{\theta}_2,\cdots,\boldsymbol{\theta}_K\}$, model merging aims to combine $K$ fine-tuned models into one single model simultaneously performing on $\{T_1,T_2,\cdots,T_K\}$ without post-training~\cite{method_p1_1,method_p1_2}.
Task vector~\cite{ilharco2023editing,yang2024adamerging} is a key element in merging method, which could enhance the base model's ability or enable the model to handle other tasks. Specifically, for task $T_i$, the task vector $\boldsymbol\tau_i\in \mathbb{R}^D$ is defined as the vector obtained by subtracting the SFT weights $\boldsymbol{\theta}_i$ from the base model weight
$\boldsymbol{\theta}_{\rm{base}}$, \emph{i.e.}, $\boldsymbol\tau_i=\boldsymbol{\theta}_i-\boldsymbol{\theta}_{\rm{base}}$. The parameters of merged model could be denoted as $\boldsymbol{\theta}_m=\boldsymbol{\theta}_{\rm{base}}+\sum_i \lambda_i\boldsymbol{\tau}_i$, where $\lambda_i$ is the scaling factor measuring the importance of task vectors and $D$ denotes the dimensionality of the flattened weight vector. For clarity, we denote the neuron set in $\boldsymbol{\theta}_i$ as $\mathcal{N}_i$, and that in $\boldsymbol{\tau}_i$ as $\mathcal{T}_i$

\begin{figure*}[t]
            \vspace{-20pt}
            \centering
            \includegraphics[width=\textwidth]{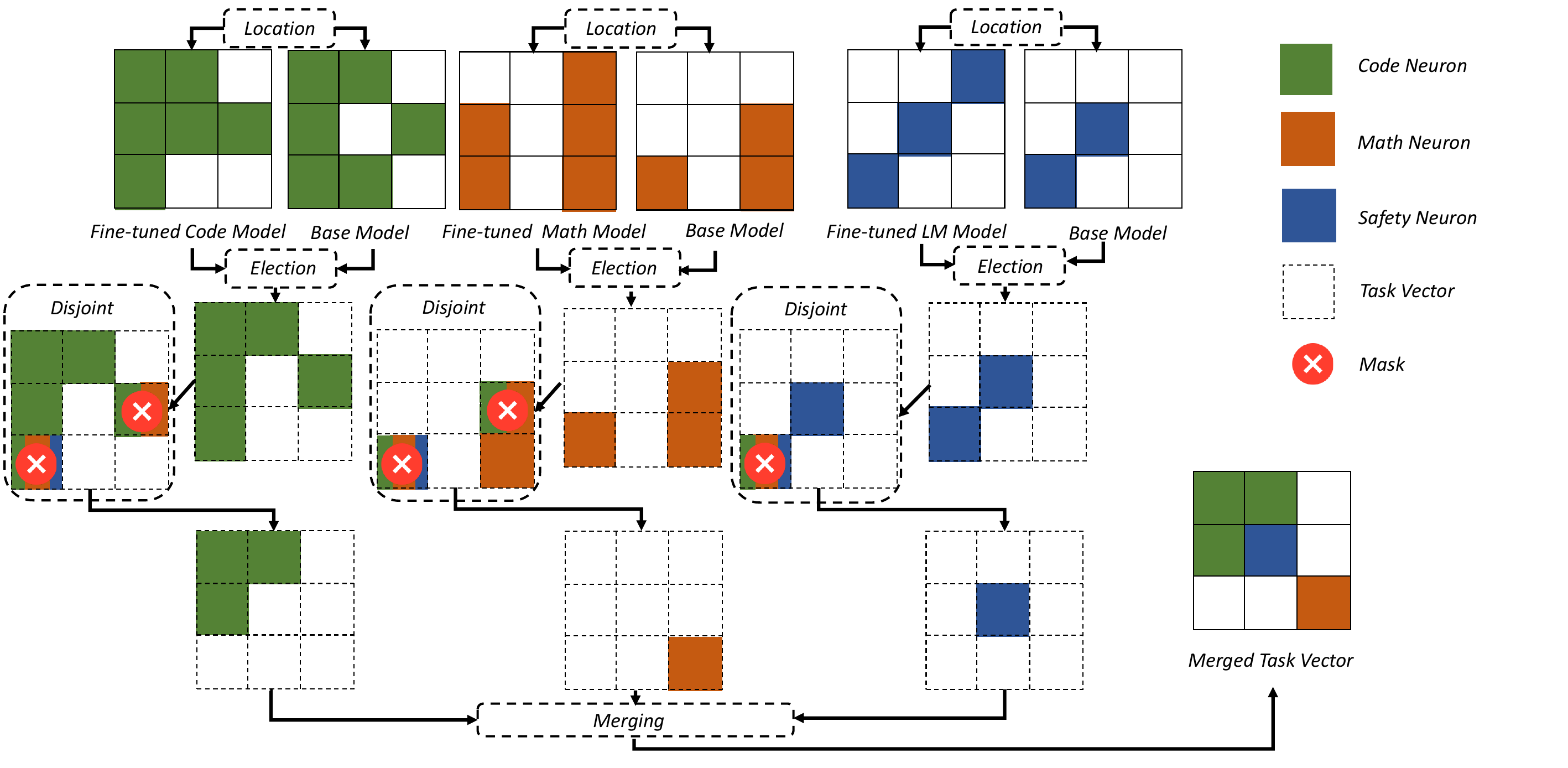}
            \vspace{-20pt}
            \caption{Overview of LED-Merging. In \textbf{location}, we identify important safety and utility neurons in base and fine-tuned models, respectively. We use different colors to represent the various neurons. After location, in \textbf{election}, we select neurons scoring highly in both two models in the election step as safety and utility-related neurons in task vectors. Subsequently, we \textbf{disjoint} these important neurons and construct masks, in which isolating the duplicated important neurons across all task vectors. Finally, we combine them into one merged task vector.}
     \label{fig:method}
    \vspace{-10pt}
\end{figure*}

\paragraph{LED-Merging: Location, Election, and Disjoint Merging.}
Previous studies \cite{modelstock, ilharco2023editing, tiesmerging} fail to accurately identify safety-related neurons in task vectors with a single magnitude score, namely \textit{neuron misidentification}. Meanwhile, there exists an interference between safety-related and utility-related task vector neurons during the merging process, namely \textit{neuron interference}. To address neuron misidentification, we first \textit{locate} important neurons both in the base and fine-tuned models, and then \textit{elect} neurons from the task vector considering these two scores together. Subsequently, to mitigate the interference, we introduce a \textit{disjoint} step, isolating these important neurons so that they influence different base neurons. The whole process is illustrated in Figure~\ref{fig:method}.

\begin{algorithm}[!ht]
    \caption{LED-Merging}
    \label{alg1}
    \begin{algorithmic}[1]
        \REQUIRE  base model parameters $\boldsymbol{\theta}_{\rm{base}}$, SFT models' parameters $\{\boldsymbol{\theta}_{i}\mid i\in [K]\}$, mask ratios \{$r_{i} \mid i\in [K]\}$, scaling factors $\{\lambda_i\mid i\in[K]\}$, location datasets $\{\mathcal{X}_{i}\mid i\in[K]\}$
        \ENSURE merged parameter $\boldsymbol{\theta}_{m}$
        \STATE $\boldsymbol{\theta}_{m}\leftarrow \boldsymbol{\theta}_{\rm{base}}$
        \FOR{$i\in [K]$}
        \STATE $I(\boldsymbol{\theta}_i)=\mathbb{E}_{x\sim \mathcal{X}_i}|\boldsymbol{\theta}_{i}\odot \nabla_{\boldsymbol{\theta}_i}\mathcal{L}(x)|$
        \STATE $I(\boldsymbol{\theta}_{\rm{base}})=\mathbb{E}_{x\sim \mathcal{X}_i}|\boldsymbol{\theta}_{\rm{base}}\odot \nabla_{\boldsymbol{\theta}_{\rm{base}}}\mathcal{L}(x)|$
        
        \STATE calculate $\mathcal{T}^{r_i}_{i}$ following Equation \ref{vote}

        \ENDFOR  
        \FOR{$i\in [K]$}
        
        \STATE calculate $\text{Disjoint}(\mathcal{T}_i^{r_i})$ use Equation~\ref{disjoint_safety}
        \STATE $\boldsymbol{m}_i \leftarrow \boldsymbol{0}$
        \FOR{$d\in \text{Disjoint}(\mathcal{T}_i^{r_i})$}
        \STATE $\boldsymbol{m}_{i,d}=1$
        \ENDFOR
        \STATE $\boldsymbol\tau_i=\boldsymbol{\theta}_i-\boldsymbol{\theta}_{\rm{base}}$
        \STATE $\boldsymbol{\theta}_{m}\leftarrow \boldsymbol{\theta}_{m}+\lambda_i \boldsymbol{\tau}_i\odot \boldsymbol{m}_{i}$
        \ENDFOR
    \end{algorithmic}
\end{algorithm}
In the location and election step, we consider the importance score from base and fine-tuned models simultaneously to locate task-specific neurons. In this way, it is more accurate than relying on the magnitude score alone because task-specific neurons with high importance score in the fine-tuned model may not necessarily score high in the base model, and vice versa.

{\textbf{Location}}.  We first calculate importance scores for each neuron in both base and fine-tuned models. Given a location dataset $\mathcal{X}_i=\{(x,y)_k\}$, where $x$ is the question and $y$ is the answer, we calculate the importance scores for the weight $\boldsymbol{\theta}_i\in\mathbb{R}^D$ in any  layer as follows~\cite{snip,spareseGPT,sun2024a}:
\begin{equation}
    I(\boldsymbol{\theta}_i)=\mathbb{E}_{x\sim \mathcal{X}_i}[\boldsymbol{\theta}_i\odot \nabla _{\boldsymbol{\theta}_i}\mathcal{L}(x)],
    \label{location}
\end{equation}
which $\mathcal{L}(x)=-\log p(y\mid x)$ is the conditional negative log-likelihood loss. We choose the SNIP score~\cite{snip} because it balances computational efficiency and performance~\cite{cq}. Please refer to Sec.~\ref{sec:ablation} for the comparison between different location methods. After computing importance scores, we choose top-$r_i$ neurons as the important neuron subset $\mathcal{N}_{i}^{r_i}$ from $I(\boldsymbol{\theta}_i)$.
 

{\textbf{Election}}. A natural question is how to select important neurons in the task vector $\boldsymbol{\tau}_i$ based on $I(\boldsymbol{\theta}_{\rm{base}})$ and $I(\boldsymbol{\theta}_{i})$. The important neurons in the base model may be different from neurons in the fine-tuned model. Therefore, we introduce the following election strategy to select neurons with high scores in both base and fine-tuned models:
\begin{equation}
    \mathcal{T}_i^{r_i}=\mathcal{N}_i^{r_i}\cap \mathcal{N}_{\rm{base}}^{r_i}.
    \label{vote}
\end{equation}
\emph{Remark}. We compare different choosing methods, including scoring low or high in base or fine-tuned model in Section~\ref{sec:ablation} and find that Equation \ref{vote} achieves the best performance.

{\textbf{Disjoint}}. As important neurons from different task vectors may conflict with each other at the same position, we use the set difference to disjoint the neurons from others to prevent interference:
\begin{equation}
    \text{Disjoint}(\mathcal{T}^{r_i}_{i})=\mathcal{T}^{r_i}_{i}-\mathop{\cup}\limits_{{J}\subsetneqq [K],|J|\geq 2}\mathop{\cap}\limits_{j\in {J}}\mathcal{T}^{r_j}_{j}.
    \label{disjoint_safety}
\end{equation}
This operation removes all neurons that appear in at least two tasks' elected sets, ensuring that $\text{Disjoint}(\mathcal{T}_i)$ only contains neurons uniquely attributed to task $i$.
Next, we construct a mask $\boldsymbol{m}_i\in\mathbb{R}^D$ to implement disjoint in the merging process. Specifically, this mask $\boldsymbol{m}_i$ is used to select neurons from $\mathcal{T}_i$. The mask ratio is $r_i$, where $r\in(0,1]$. The mask $\boldsymbol{m}_i$ can be derived from:
\begin{equation}
    \boldsymbol{m}_{i,d}=\begin{aligned} &\left\{ \begin{array}{ll} 1, & \text{if } d\in \text{Disjoint}(\mathcal{T}_{i}^{r_i}), \\ 0, & \text{otherwise}. \end{array} \right. \end{aligned}
    \label{mask_safety}
\end{equation}

{\textbf{Merging}}. The final
merged task vector $\boldsymbol{\tau}_m$ is as follows:
\begin{equation}
    \boldsymbol{\tau}_m= \sum_i \lambda_i\boldsymbol{\tau}_{i}\odot\boldsymbol{m}_i.
    \label{merged_task_vector}
\end{equation}
We summarize the workflow in Algorithm \ref{alg1}.

\section{Experiments}

\begin{table*}[!htbp]
\centering
\vspace{-10pt}
\caption{Performance of merging Llama3-8B-Instruct (LM), MAmmoTH2-8B-Plus (Math), and Replete-Coder-Llama3-8B (Code) on all the datasets. The best and second-best results are marked in \textbf{bold} and \underline{underlined} fonts. $\ast$: The merged model fails to provide structured response.}
\vspace{-10pt}
\label{tab:llms_merging_llama3}
\resizebox{\textwidth}{!}
{
\begin{tabular}{c|ccc|cc|cc|cc}
\toprule
\multirow{2}{*}{\begin{tabular}[c]{@{}c@{}}\textbf{Merging} \\ \textbf{Methods}\end{tabular}} & \multicolumn{3}{c|}{\begin{tabular}[c]{@{}c@{}}\textbf{Models}\end{tabular}} & \multicolumn{2}{c|}{\begin{tabular}[c]{@{}c@{}}\textbf{Safety Alignment}\end{tabular}} & 
\multicolumn{2}{c|}{\begin{tabular}[c]{@{}c@{}}\textbf{Mathematical} \\ \textbf{Reasoning}\end{tabular}} & \multicolumn{2}{c}{\begin{tabular}[c]{@{}c@{}}\textbf{Code Generating}\end{tabular}} \\ \cmidrule{2-10} 
                                                                            & \textbf{LM} & \textbf{Math} & \textbf{Code} &                             \textbf{HarmBench}$\downarrow$                                                     & \textbf{SORRY-Bench}$\downarrow$                                                       & \textbf{GSM8K}$\uparrow$                                          & \textbf{MATH}$\uparrow$                                  & \textbf{MBPP}$\uparrow$                              & \textbf{HumanEvalPack}$\uparrow$                                  \\ \midrule
\multirow{3}{*}{\begin{tabular}[c]{@{}c@{}}w/o Merging \\ \end{tabular}}    & \textcolor{gray}{\usym{2714}} & &                     & 21.50                                                                               & 18.67                                                            & 81.05                                        & 24.56                                 & 1.00                                  & 3.65                                 \\  
                                                                            & &\textcolor{gray}{\usym{2714}} &                  & 42.00                                                                               & 50.60                  & 79.00                                          & \textbf{36.72}                                 & /                                      & /                                    \\  
                                                                            & & &\textcolor{gray}{\usym{2714}}                    & 61.25                                                                                & 90.40                                                                & /                                              & /                                     & 33.60                                  & \textbf{42.68}                                 \\ \midrule
\multirow{3}{*}{\begin{tabular}[c]{@{}c@{}}Model Stock\end{tabular}} &\textcolor{gray}{\usym{2714}} &\textcolor{gray}{\usym{2714}} &               & 36.00                                                                               & 39.55                                                            & 59.67                                  & 16.64                                 & /                                  & /                                                                  \\ 
                                                                            & \textcolor{gray}{\usym{2714}} & & \textcolor{gray}{\usym{2714}}             & \underline{17.25}                                                                               & 12.67                                                   & /                                             & /                                      & 47.00                                   & \underline{39.02}                                                    \\  
                                                                            &\textcolor{gray}{\usym{2714}} &\textcolor{gray}{\usym{2714}} &\textcolor{gray}{\usym{2714}}           & 23.25                                           & 17.78                                        & 52.92                                          &  15.22                                 &  47.80                                   & 36.59                                                                  \\ 
                                                                                                              \midrule
\multirow{3}{*}{\begin{tabular}[c]{@{}c@{}}Breadcrumbs\end{tabular}} &\textcolor{gray}{\usym{2714}} &\textcolor{gray}{\usym{2714}} &              & 33.00                                                                               & 35.78                                                            & $\ast$                                   & $\ast$                                & /                                  & /                                                                  \\  
                                                                            & \textcolor{gray}{\usym{2714}} & &\textcolor{gray}{\usym{2714}}             & 39.50                                                                              & 36.89                                                   & /                                             & /                                      & \textbf{53.40}                                   & 36.58                                                     \\  
                                                                            & \textcolor{gray}{\usym{2714}} &\textcolor{gray}{\usym{2714}} &\textcolor{gray}{\usym{2714}}           & 38.25                                           & 40.44                                         & $\ast$                                           & $\ast$                                  & 49.40                                    & 36.59                                                                   \\  
                                                                                                              \midrule
\multirow{3}{*}{\begin{tabular}[c]{@{}c@{}}Task \\ Arithmetic\end{tabular}} & \textcolor{gray}{\usym{2714}} &\textcolor{gray}{\usym{2714}} &              & 26.50                                                                               & 28.89                                                            & 54.59                                  & 16.77                                 & /                                  & /                                                                  \\  
                                                                            & \textcolor{gray}{\usym{2714}} & & \textcolor{gray}{\usym{2714}}            & 38.00                                                                               & 31.11                                                   & /                                             & /                                      & 37.8                                   & 18.90                                                    \\  
                                                                            & \textcolor{gray}{\usym{2714}} &\textcolor{gray}{\usym{2714}} &\textcolor{gray}{\usym{2714}}           & 32.00                                           & 38.44                                        & 13.12                                           & 9.92                                  & 21.8                                    & 9.15                                                                   \\  
                                                                                                              \midrule
\multirow{3}{*}{\begin{tabular}[c]{@{}c@{}}Ties- \\ Merging\end{tabular}} &\textcolor{gray}{\usym{2714}}  &\textcolor{gray}{\usym{2714}} &             & 35.75                                                                               & 37.11                                                             & 55.37                                  & 17.45                                 & /                                   & /                                                                  \\  
                                                                            &\textcolor{gray}{\usym{2714}} & &\textcolor{gray}{\usym{2714}}             & 45.00                                                                                & 46.44                                                    & /                                              & /                                      & 41.60                                   & 33.53                                                    \\             &\textcolor{gray}{\usym{2714}} &\textcolor{gray}{\usym{2714}} &\textcolor{gray}{\usym{2714}}           & 41.25                                                                                & 46.44                                                                 & 53.01                                           & 16.72                                  & \underline{50.20}                                   & 30.34                                                                  \\  
                                                                            \midrule
                                                                            
\multirow{3}{*}{\begin{tabular}[c]{@{}c@{}}LED- \\ Merging(Ours)\end{tabular}} &\textcolor{gray}{\usym{2714}} &\textcolor{gray}{\usym{2714}} &              & 21.00                                                                               & 11.33                                                            & 49.89                                  & 16.12                                 & /                                  & /                                                                \\  
                                                                            &\textcolor{gray}{\usym{2714}} & &\textcolor{gray}{\usym{2714}}             & \textbf{14.75}                                                                               & \textbf{10.22}                                                    & /                                              & /                                      & 47.2                                & 37.80                                                   \\             & \textcolor{gray}{\usym{2714}} &\textcolor{gray}{\usym{2714}} &\textcolor{gray}{\usym{2714}}           & 20.75                                                                               & \underline{10.44}                                                                & 52.39                                         & 15.08                             & 44.6                                  & 36.59                                                                  \\  
                                                                            \bottomrule                                                                        
\end{tabular}
}
\vspace{-8pt}
\end{table*}

\subsection{Experimental Setup}
\label{sec:exp_setup}
\noindent\textbf{Baselines.} We compare LED-Merging with multiple merging methods: \textbf{Model Stock}~\cite{modelstock}, \textbf{Model Breadcrumbs}~\cite{breadcrumbs}, \textbf{Task Arithmetic}~\cite{ilharco2023editing}, \textbf{Ties-Merging}~\cite{tiesmerging}. Please see Appendix~\ref{sec:appendix_baselines} for more discussions.

\noindent\textbf{Datasets\&Metrics.}
We assess safety-utility trade-offs through three pillars: (1) Safety via HarmBench~\cite{harmbench} and SORRY-Bench~\cite{xie2024sorrybench} (Attack Success Rate, ASR$\downarrow$). (2) Mathematical reasoning using GSM8K~\cite{gsm8k} and MATH~\cite{hendrycksmath2021} (Accuracy$\uparrow$ with chain-of-thought. (3) Code Generation evaluated by MBPP~\cite{mbpp} and HumanEvalPack~\cite{humanevalpack} (Pass@1$\uparrow$). More detailed task descriptions and verification protocols appear in Appendix~\ref{sec:appendix_dataset}

\noindent\textbf{Models.} 
We evaluate three model families: (1) Llama-3 (8B base/instruct/math/code variants), (2) Wizard-LM (13B base/instruct/math/code), and (3) Mistral (7B base/instruct/math). All models use base architectures paired with safety-aligned or task-specialized versions. Further Details of SFT models are in Appendix~\ref{sec:sft-models-info}.

\noindent\textbf{Implementation Details.} Following~\cite{dare,xu2024wizardlm,hendrycksmath2021}, inference is implementated by vLLM~\cite{kwon2023efficient}. We use grid search to obtain optimal hyperparameters for both baselines and optimal mask ratios for our LED-Merging, recommended hyperparameters are listed in Appendix~\ref{sec:hyperparam-baselines}. 

\subsection{Hyperparameter Analysis}
LED-Merging’s robustness stems from its ability to balance safety-utility trade-offs through two key hyperparameters: \textit{mask ratios \(r_i\)} controlling neuron retention, and \textit{scaling factors \(\lambda_i\)} governing task vector contributions. Experiments on Mistral-7B reveal distinct design principles.

\begin{figure}[!ht]
            \centering
            \includegraphics[width=\linewidth]{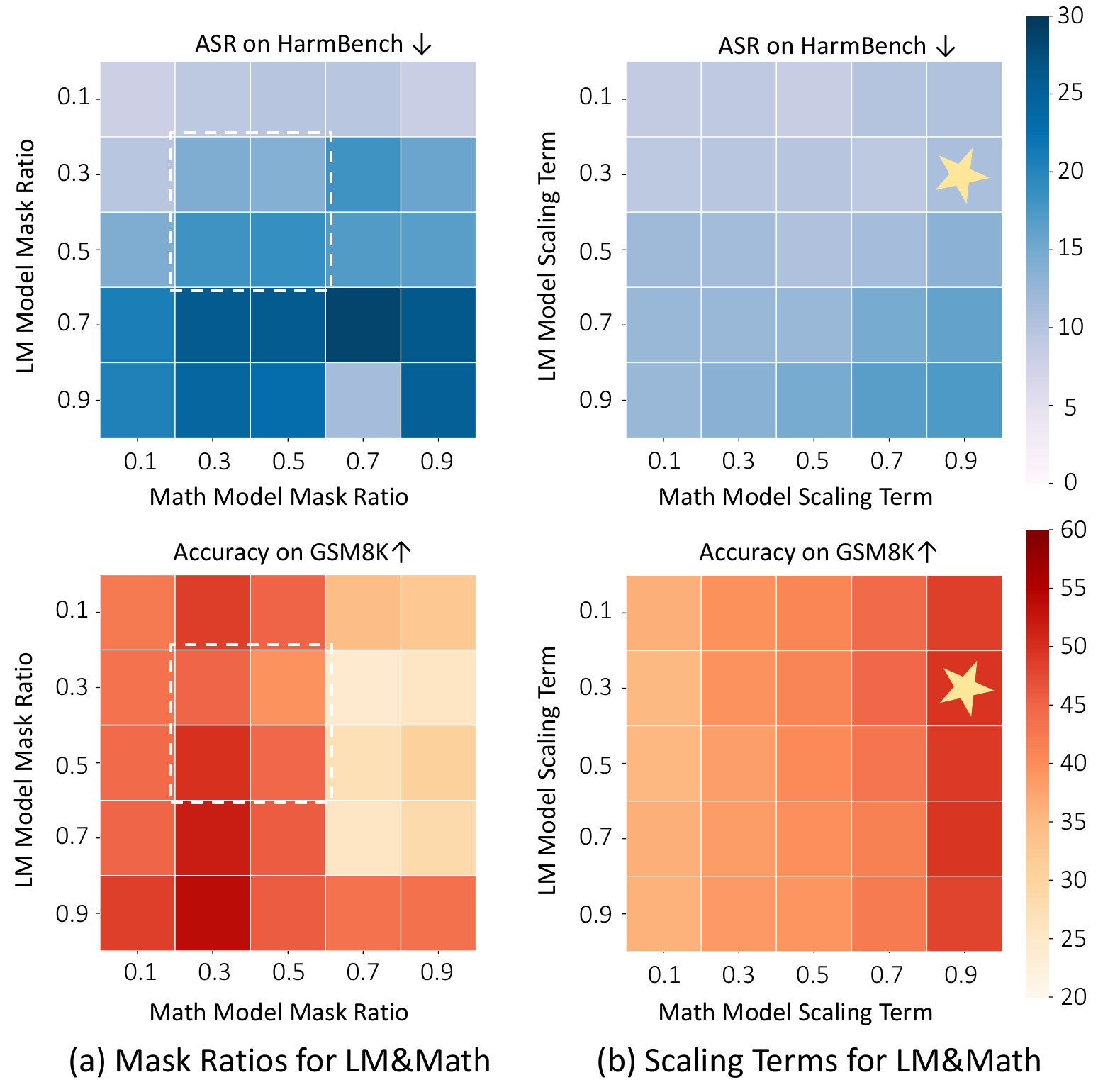} 
            \caption{Safety-utility trade-offs under varying hyperparameters. 
            (a)~\textbf{Mask ratios}: \textcolor{darkblue}{Blue} means better safety alignment (lower ASR), while \textcolor{orange}{orange} means better math ability (higher Accuracy). The Pareto frontier (white dashed line) reveals optimal ratios (0.3–0.5) balancing both metrics.
            (b)~\textbf{Scaling terms}: demonstrates safety degradation with maintained utility performance. Star markers denote configurations achieving >90\% safety preservation with <5\% utility loss.}
     \label{fig:sensitive}
     \vspace{-20pt}
\end{figure}

\noindent\textbf{Mask ratio dynamics.}
As shown in Fig.~\ref{fig:sensitive}\textcolor{darkblue}{a}, varying \(r_{\text{LM}}\)(safety) and \(r_{\text{Math}}\)(utility) reveals three critical regimes. In \textbf{safety-centric mode}~($r_{\text{i}}\leq 0.3$), prioritizing base model neurons($r_{\text{LM}}=0.1$) minimizes ASR to 7.75\%, but suppresses math capabilities (42.38\% accuracy). Conversely, \textbf{utility-centric mode}($r_{\text{i}}\geq 0.5$) maximizes accuracy on GSM8K to 53.68\% by retaining task-specific neurons, yet compromises safety~(ASR > 25\%). The \textbf{Pareto-optimal regime}~($r_{i}=0.3-0.5$, labeled by white dashed line) strikes a balance. When $r_{\text{LM}}, r_{\text{Math}}=0.5$, 18.75\% ASR and 44.81\% accuracy are achieved through spatially disjoint neuron updates, confirming that moderate ratios maximize conflict-free parameter fusion.

\noindent\textbf{Scaling factor trade-offs.}
Scaling factors \(\lambda_i\) dictate the dominance hierarchy between safety and utility gradients, shown in Fig.~\ref{fig:sensitive}\textcolor{darkblue}{b}. Amplifying safety contributions~($\lambda_{\text{LM}} \geq 0.7$) suppresses harmful behaviors but over-penalizes mathematical ability. 
Prioritizing utility boosts accuracy on GSM8K, yet reintroduces safety risks. The equilibrium configuration, labeled by a star marker, achieves 11\% ASR and 49.66\% accuracy, demonstrating a balanced task coexistence.

\subsection{Main Results}
\begin{table}[ht]
\centering
\caption{Performance of merging Mistral-Instruct-7B (LM) and MetaMath-Mistral-7B (Math) on all the datasets. The best and second-best results are marked in \textbf{bold} and \underline{underlined} fonts. $\ast$: The instruction following ability of LLMs is destroyed, discussed in Sec.~\ref{sec:discussion_impair}.}
\label{tab:llms_merging_mistral}
\vspace{-10pt}
\resizebox{\linewidth}{!}
{
\begin{tabular}{c|cc|cc|cc}
\toprule
\multirow{1}{*}{\begin{tabular}[c]{@{}c@{}}\textbf{Merging} \\ \textbf{Methods}\end{tabular}} & \multicolumn{2}{c|}{\begin{tabular}[c]{@{}c@{}}\textbf{Models}\end{tabular}} & \multicolumn{2}{c|}{\begin{tabular}[c]{@{}c@{}}\textbf{Safety}\end{tabular}} & 
\multicolumn{2}{c}{\begin{tabular}[c]{@{}c@{}}\textbf{Mathematical} \\ \textbf{Reasoning}\end{tabular}} \\ \cmidrule{2-7} 
                                                                            & \textbf{LM}   & \textbf{Math}                      &                             \textbf{HarmBench}$\downarrow$                                                     & \textbf{SORRY-Bench}$\downarrow$                                                       & \textbf{GSM8K}$\uparrow$                                          & \textbf{MATH}$\uparrow$                                                                    \\ \midrule
\multirow{2}{*}{\begin{tabular}[c]{@{}c@{}}w/o Merging \\ \end{tabular}}    &  \textcolor{gray}{\usym{2714}} &                     & 54.75                                                                               & 53.11                                                           & 50.19                                        & 9.74                                  \\  
                                                                            & &  \textcolor{gray}{\usym{2714}}                   & 46.00                                                                               & 69.56                  & \textbf{75.36}                                          & \textbf{27.32}                                \\ \midrule
\multirow{1}{*}{\begin{tabular}[c]{@{}c@{}}Model Stock\end{tabular}} &\textcolor{gray}{\usym{2714}} &\textcolor{gray}{\usym{2714}}               & $\ast$                                                                               & $\ast$                                                           & 35.41                                & 9.52                               \\ 
\multirow{1}{*}{\begin{tabular}[c]{@{}c@{}}Breadcrumbs\end{tabular}} &\textcolor{gray}{\usym{2714}} &\textcolor{gray}{\usym{2714}}               & 63.75                                                                               & 70.89                                                            & \underline{63.99}                                & \underline{15.96}                               \\ 
\multirow{1}{*}{\begin{tabular}[c]{@{}c@{}}Task Arithmetic\end{tabular}} &\textcolor{gray}{\usym{2714}} & \textcolor{gray}{\usym{2714}}              & 55.75                                                                               & 69.55                                                           & 60.88                                  & 15.14                                 \\ 
\multirow{1}{*}{\begin{tabular}[c]{@{}c@{}}Ties-Merging\end{tabular}} &\textcolor{gray}{\usym{2714}} & \textcolor{gray}{\usym{2714}}              & 62.00                                                                               & 79.78                                                             & 59.21                                  & 13.50                                 \\
\multirow{1}{*}{\begin{tabular}[c]{@{}c@{}}LED-Merging(Ours)\end{tabular}} &\textcolor{gray}{\usym{2714}} &\textcolor{gray}{\usym{2714}}               & \underline{16.00}                                                                               & \underline{24.22}                                                            & 50.34                                  & 14.20                                 \\ \bottomrule                                                                        
\end{tabular}
}
\vspace{-10pt}
\end{table}

\noindent\textbf{LED-Merging presents superior safety capacity.}
LED-Merging achieves SOTA safety performance across all evaluated benchmarks, surpassing both existing merging methods and even the original safety-aligned models.
On HarmBench, merging safety-aligned and code-specialized models for Llama3-8B reduces the ASR to 14.75\%, a \textbf{75.9\%} improvement over the standalone code model (ASR=61.25\%) and a 31.4\% enhancement compared to the original LM model (ASR=21.50\%). This indicates that LED-Merging not only mitigates conflicts but actively strengthens safety through gradient-informed neuron election.
Similar trends hold for Mistral-7B, where merging safety and math models achieves ASR=16\%, outperforming Task Arithmetic (ASR=55.75\%) and Ties-Merging (ASR=62\%) while surpassing the original Mistral-7B-Instruct (ASR=54.75\%) by \textbf{70.8\%}. For larger models like Llama2-13B, merging multiple specific fine-tuned models maintains an exceptionally low ASR=4\%, significantly better than both baselines (Task Arithmetic: 35\%) and the standalone safety model (28.25\%), proving its capacity to resolve cross-task interference at scale.

\begin{table*}[!ht]
\centering
\vspace{-10pt}
\caption{Performance of merging WizardLM-13B (LM), WizardMath-13B (Math), and LLama-2-13B-Code-Alpaca (Code) on all the datasets. The best and second-best results are marked in \textbf{bold} and \underline{underlined} fonts. $\ast$: The instruction following ability of LLMs is impaired, discussed in Sec.~\ref{sec:discussion_impair}.}
\vspace{-10pt}
\label{tab:llms_merging_llama2}
\resizebox{\textwidth}{!}
{
\begin{tabular}{c|ccc|cc|cc|cc}
\toprule
\multirow{2}{*}{\begin{tabular}[c]{@{}c@{}}\textbf{Merging} \\ \textbf{Methods}\end{tabular}} & \multicolumn{3}{c|}{\begin{tabular}[c]{@{}c@{}}\textbf{Models}\end{tabular}} & \multicolumn{2}{c|}{\begin{tabular}[c]{@{}c@{}}\textbf{Safety Alignment}\end{tabular}} & 
\multicolumn{2}{c|}{\begin{tabular}[c]{@{}c@{}}\textbf{Mathematical} \\ \textbf{Reasoning}\end{tabular}} & \multicolumn{2}{c}{\begin{tabular}[c]{@{}c@{}}\textbf{Code Generating}\end{tabular}} \\ \cmidrule{2-10} 
                                                                            & \textbf{LM} & \textbf{Math} & \textbf{Code} &                           \textbf{HarmBench}$\downarrow$                                                     & \textbf{SORRY-Bench}$\downarrow$                                                       & \textbf{GSM8K}$\uparrow$                                          & \textbf{MATH}$\uparrow$                                  & \textbf{MBPP}$\uparrow$                              & \textbf{HumanEvalPack}$\uparrow$ \\ \midrule
\multirow{3}{*}{\begin{tabular}[c]{@{}c@{}}w/o Merging \\ \end{tabular}}    & \textcolor{gray}{\usym{2714}} & &                     & 28.25                                                                               & 32.44                                                            & 52.16                                        & \underline{7.42}                                  & 31.40                                  & 26.22                                 \\  
                                                                            & &\textcolor{gray}{\usym{2714}} &                   & 45.75                                                                               & 56.67                  & 64.22                                          & \textbf{14.02}                                 & /                                      & /                                    \\  
                                                                            & & &\textcolor{gray}{\usym{2714}}                   & 15.50                                                                                & \underline{12.44}                                                                & /                                              & /                                     &  22.82                                  & 22.56                                 \\ \midrule
\multirow{3}{*}{\begin{tabular}[c]{@{}c@{}}Model Stock\end{tabular}} & \textcolor{gray}{\usym{2714}} &\textcolor{gray}{\usym{2714}} &              & 20.00                                                                               & 24.67                                                            & 26.99                                  & 5.26                                 & /                                  & /                                                                  \\  
                                                                            & \textcolor{gray}{\usym{2714}} & &\textcolor{gray}{\usym{2714}}             & $\ast$                                                                               & $\ast$                                                   & /                                             & /                                      & 14.00                                   & 14.02                                                    \\  
                                                                            & \textcolor{gray}{\usym{2714}} &\textcolor{gray}{\usym{2714}} &\textcolor{gray}{\usym{2714}}           & $\ast$                                           & $\ast$                                        & 25.17                                           & 1.36                                   & 6.20                                    & 15.85                                                                   \\  
                                                                                                              \midrule 
\multirow{3}{*}{\begin{tabular}[c]{@{}c@{}}Breadcrumbs\end{tabular}} & \textcolor{gray}{\usym{2714}}  &\textcolor{gray}{\usym{2714}} &             & 47.25                                                                               & 38.22                                                            & 63.76                                  &  0.36                                & /                                  & /                                                                  \\  
                                                                            & \textcolor{gray}{\usym{2714}} & &\textcolor{gray}{\usym{2714}}             & 37.75                                                                               & 40.44                                                   & /                                             & /                                      & 32.00                                   & \textbf{32.93}                                                    \\  
                                                                            & \textcolor{gray}{\usym{2714}} &\textcolor{gray}{\usym{2714}} &\textcolor{gray}{\usym{2714}}           & 44.75                                           & 40.44                                        & \textbf{65.81}                                           & 1.66                                  & 28.40                                    & 20.73                                                                   \\  
                                                                                                              \midrule 
\multirow{3}{*}{\begin{tabular}[c]{@{}c@{}}Task \\ Arithmetic\end{tabular}} & \textcolor{gray}{\usym{2714}} &\textcolor{gray}{\usym{2714}} &              & 31.75                                                                                & 34.67                                                             & 61.94                                  & 2.22                                  & /                                  & /                                                                  \\  
                                                                            & \textcolor{gray}{\usym{2714}} & &\textcolor{gray}{\usym{2714}}             & 37.00                                                                                & 35.56                                                   & /                                             & /                                      & \underline{33.20}                                   & \underline{31.09}                                                    \\  
                                                                            & \textcolor{gray}{\usym{2714}} &\textcolor{gray}{\usym{2714}} &\textcolor{gray}{\usym{2714}}           & 35.00                                           & 37.78                                        & 61.56                                           & 5.00                                  & 25.80                                    & 16.46                                                                   \\  
                                                                                                              \midrule
\multirow{3}{*}{\begin{tabular}[c]{@{}c@{}}Ties- \\ Merging\end{tabular}} & \textcolor{gray}{\usym{2714}} &\textcolor{gray}{\usym{2714}} &              & 39.50                                                                               & 40.00                                                            & \underline{65.58}                                  & 4.58                                 & /                                  & /                                                                  \\  
                                                                            & \textcolor{gray}{\usym{2714}} &\textcolor{gray}{\usym{2714}} &             & 37.00                                                                               & 43.56                                                   & /                                             & /                                      & 33.00                                   & 28.66                                                    \\  
                                                                            & \textcolor{gray}{\usym{2714}} &\textcolor{gray}{\usym{2714}} &\textcolor{gray}{\usym{2714}}           & 38.25                                           & 40.22                                        &  62.92                                          & 0.74                                  & 31.40                                    & 26.83                                                                   \\  
                                                                                                              \midrule
\multirow{3}{*}{\begin{tabular}[c]{@{}c@{}}LED- \\ Merging~(Ours)\end{tabular}} &\textcolor{gray}{\usym{2714}} &\textcolor{gray}{\usym{2714}} &              & \underline{13.75}                                                                               & \textbf{11.78}                                                            & 43.97                                  & 4.10                                 & /                                  & /                                                                  \\  
                                                                            & \textcolor{gray}{\usym{2714}} & &\textcolor{gray}{\usym{2714}}             & 14.75                                                                               & 15.11                                                   & /                                             & /                                      & \textbf{33.80}                                   & 15.85                                                    \\  
                                                                            &\textcolor{gray}{\usym{2714}}  &\textcolor{gray}{\usym{2714}} &\textcolor{gray}{\usym{2714}}           & \textbf{4.00}                                           & 17.11                                        & 46.40                                           & 5.84                                  & 23.80                                    & 17.68                                                                   \\  
                                                                                                              \bottomrule                                                                        
\end{tabular}
}
\vspace{-10pt}
\end{table*}

\begin{table}[!ht]
\centering
\caption{Ablation Study. Experiments are conducted on Mistral-7B series models. $\ast$ represents LLM's instruction following ability is impaired.}
\vspace{-10pt}
\resizebox{\linewidth}{!}{
\begin{tabular}{c|c|cc|cc}
\toprule
\multirow{2}{*}{\begin{tabular}[c]{@{}c@{}}\textbf{Ablation} \\ \textbf{Part}\end{tabular}} & \multirow{2}{*}{\begin{tabular}[c]{@{}c@{}}\textbf{Alternative} \\ \textbf{Methods}\end{tabular}} & \multicolumn{2}{c|}{\begin{tabular}[c]{@{}c@{}}\textbf{Safety}\end{tabular}} & 
\multicolumn{2}{c}{\begin{tabular}[c]{@{}c@{}}\textbf{Mathematical} \\ \textbf{Reasoning}\end{tabular}} \\ \cmidrule{3-6}
&                         &                             \textbf{HarmBench}$\downarrow$                                                     & \textbf{SORRY-Bench}$\downarrow$                                                       & \textbf{GSM8K}$\uparrow$                                          & \textbf{MATH}$\uparrow$                                   \\ \midrule

\multirow{3}{*}{Location} & Random                   & $\ast$             & $\ast$              & 25.58              & 8.66             \\ 
                         & Wanda       & $\ast$             & $\ast$               & 39.58              & 11.37             \\  
                        & SNIP               & 16.00        & 24.22          & 50.34         & 14.20        \\ \midrule

\multirow{3}{*}{Election} 
                   & 01       &  58.00            & 83.77               & 54.13              & 13.12             \\ 
                   & 10                & 35.25             & 47.33               & 50.64              & 13.30             \\ 
                    & 11               & 16.00        & 24.22          & 50.34         & 14.20        \\ \midrule

\multirow{2}{*}{Disjoint}          & \textcolor{gray}{\usym{2717}}      & 63.00             & 85.33               & 72.93              & 23.18             \\
           & \textcolor{gray}{\usym{2714}}               & 16.00        & 24.22          & 50.34         & 14.20        \\ \bottomrule
\end{tabular}
}
\vspace{-10pt}
\label{tab:ablation}
\end{table}

\noindent\textbf{LED-Merging preserves utility performance with safety alignment.}
Beyond superior safety-alignment performance, LED-Merging maintains comparable utility performance to specialized models and merging baselines across mathematical reasoning and code generation tasks. When merging Llama3-8B's safety-aligned model with its math-specialized counterpart, our method retains 52.39\% accuracy on GSM8K—significantly outperforming Task Arithmetic (13.12\%) and closely matching Ties-Merging (53.01\%)—while preserving 66.3\% of the math-specialized model's capability (79.00\%). Similar advantages emerge in code generation, where merging safety and code models yields 47.2\% MBPP Pass@1 (40.2\% higher than the code-specialized model's 33.6\%), demonstrating effective preservation of specialized capabilities. 
Crucially, for safety, math, and code multi-task merging scenarios, LED-Merging sustains 52.39\% GSM8K accuracy and 44.6\% MBPP Pass@1, surpassing Task Arithmetic by 39.3\% and 22.8\% respectively, while reducing ASR by 35.2\% (20.75\% vs 32.00\%). These results validate our approach's dual capacity to isolate task-critical neurons and suppress destructive parameter conflicts.

\noindent\textbf{LED-Merging demonstrates cross-architecture robustness.}
LED-Merging demonstrates consistent effectiveness across distinct model architectures, including Llama-2, Llama-3, and Mistral families.
For Llama3-8B, merging safety-aligned and math-specialized models preserves 52.39\% GSM8K accuracy while maintaining 20.75\% ASR.
Similarly, in Mistral-7B, a model family optimized for efficiency through sliding window attention~\cite{jiang2023mistral7b}, merging safety and math models retains 50.34\% GSM8K accuracy with 16\% ASR on HarmBench, proving compatibility with diverse architectural designs.

\begin{table*}[ht]
\centering
\vspace{-10pt}
\caption{Performance across different merging methods and languages on MGSM8KInstruct. Within each language, the best and second-best results are marked in \textbf{bold} and \underline{underlined} fonts among different merging methods.}
\vspace{-10pt}
\resizebox{\linewidth}{!}
{
\begin{tabular}{c| cc| cccccccccc | c}
\toprule
\multirow{2}{*}{\begin{tabular}[c]{@{}c@{}}\textbf{Merging} \\ \textbf{Methods}\end{tabular}} & \multicolumn{2}{c|}{\begin{tabular}[c]{@{}c@{}}\textbf{Models}\end{tabular}} & \multicolumn{11}{c}{\begin{tabular}[c]{@{}c@{}}\textbf{MGSM8KInstruct~\cite{chen2023breaking}}\end{tabular}} \\
\cmidrule(lr){2-14}
& \textbf{LM} & \textbf{Math} & \textbf{En} & \textbf{Sw} & \textbf{Zh} & \textbf{Bn} & \textbf{De} & \textbf{Es} & \textbf{Fr} & \textbf{Ja} & \textbf{Ru} & \textbf{Th} & \textbf{Overall} \\
\midrule
\multirow{2}{*}{\begin{tabular}[c]{@{}c@{}}w/o Merging \\ \end{tabular}}         & \textcolor{gray}{\usym{2714}} &  & 0.476 & 0.160 & 0.348 & 0.224 & 0.348 & 0.264 & 0.348 & 0.232 & 0.344 & 0.252 & 0.299 \\
&  & \textcolor{gray}{\usym{2714}} & 0.432 & 0.028 & 0.132 & 0.128 & 0.212 & 0.308 & 0.204 & 0.100 & 0.184 & 0.128 & 0.185 \\ \midrule
Model Stock              & \textcolor{gray}{\usym{2714}} & \textcolor{gray}{\usym{2714}} & 0.180 & 0.084 & 0.176 & 0.100 & 0.124 & 0.116 & 0.132 & \underline{0.184} & 0.184 & 0.120 & 0.139 \\
Breadcrumbs              & \textcolor{gray}{\usym{2714}} & \textcolor{gray}{\usym{2714}} & 0.228 & 0.072 & 0.028 & 0.136 & 0.132 & 0.136 & 0.128 & 0.096 & 0.156 & 0.120 & 0.123 \\
Task Arithmetic          & \textcolor{gray}{\usym{2714}} & \textcolor{gray}{\usym{2714}} & 0.280 & \textbf{0.100} & 0.172 & 0.128 & 0.172 & 0.132 & 0.132 & 0.144 & 0.180 & \underline{0.140} & 0.158 \\
Ties-Merging             & \textcolor{gray}{\usym{2714}} & \textcolor{gray}{\usym{2714}} & \underline{0.296} & \underline{0.092} & \underline{0.236} & \underline{0.156} & \textbf{0.216} & \underline{0.196} & \textbf{0.184} & 0.176 & \textbf{0.232} & 0.132 & \underline{0.192} \\
LED-Merging (Ours)              & \textcolor{gray}{\usym{2714}} & \textcolor{gray}{\usym{2714}} & \textbf{0.368} & 0.076 & \textbf{0.272} & \textbf{0.248} & \underline{0.204} & \textbf{0.224} & \textbf{0.184} & \textbf{0.188} & \underline{0.212} & \textbf{0.164} & \textbf{0.213} \\
\bottomrule
\end{tabular}
}
\vspace{-10pt}
\label{tab:mgsm8k}
\end{table*}

\begin{table*}[ht]
\centering
\caption{Performance across different merging methods and languages on MSVAMP. Within each language, the best and second-best results are marked in \textbf{bold} and \underline{underlined} fonts among different merging methods.}
\vspace{-10pt}
\resizebox{\linewidth}{!}
{
\begin{tabular}{c|cc|cccccccccc|c}
\toprule
\multirow{2}{*}{\begin{tabular}[c]{@{}c@{}}\textbf{Merging} \\ \textbf{Methods}\end{tabular}} & \multicolumn{2}{c|}{\begin{tabular}[c]{@{}c@{}}\textbf{Models}\end{tabular}} & \multicolumn{11}{c}{\begin{tabular}[c]{@{}c@{}}\textbf{MSVAMP~\cite{chen2023breaking}}\end{tabular}} \\
\cmidrule(lr){2-14}
& \textbf{LM} & \textbf{Math} & \textbf{En} & \textbf{Sw} & \textbf{Zh} & \textbf{Bn} & \textbf{De} & \textbf{Es} & \textbf{Fr} & \textbf{Ja} & \textbf{Ru} & \textbf{Th} & \textbf{Overall} \\
\midrule
\multirow{2}{*}{\begin{tabular}[c]{@{}c@{}}w/o Merging \\ \end{tabular}}         & \textcolor{gray}{\usym{2714}} &  & 0.703 & 0.441 & 0.586 & 0.285 & 0.615 & 0.464 & 0.556 & 0.483 & 0.533 & 0.577 & 0.524 \\
&  & \textcolor{gray}{\usym{2714}} & 0.432 & 0.186 & 0.271 & 0.204 & 0.378 & 0.495 & 0.366 & 0.238 & 0.360 & 0.287 & 0.322 \\ \midrule
Model Stock              & \textcolor{gray}{\usym{2714}} & \textcolor{gray}{\usym{2714}} & 0.323 & 0.324 & 0.424 & 0.258 & 0.422 & 0.398 & 0.389 & 0.422 & 0.576 & \underline{0.410} & 0.395 \\
Breadcrumbs              & \textcolor{gray}{\usym{2714}} & \textcolor{gray}{\usym{2714}} & 0.367 & 0.163 & 0.175 & \underline{0.384} & 0.512 & 0.523 & 0.531 & \underline{0.439} & 0.497 & \underline{0.410} & 0.400 \\
Task Arithmetic          & \textcolor{gray}{\usym{2714}} & \textcolor{gray}{\usym{2714}} & \textbf{0.750} & 0.194 & \textbf{0.601} & 0.146 & \underline{0.580} & \textbf{0.655} & \textbf{0.604} & 0.417 & \underline{0.594} & 0.387 & \underline{0.493} \\
Ties-Merging             & \textcolor{gray}{\usym{2714}} & \textcolor{gray}{\usym{2714}} & 0.494 & \underline{0.334} & 0.400 & 0.291 & \textbf{0.450} & 0.396 & 0.401 & 0.412 & 0.495 & 0.407 & 0.408 \\
LED-Merging (Ours)       & \textcolor{gray}{\usym{2714}} & \textcolor{gray}{\usym{2714}} & \underline{0.690} & \textbf{0.416} & \underline{0.561} & \textbf{0.394} & \textbf{0.628} & \underline{0.598} & \underline{0.597} & \textbf{0.517} & \textbf{0.610} & \textbf{0.450} & \textbf{0.546} \\
\bottomrule
\end{tabular}
}
\vspace{-10pt}
\label{tab:msvamp}
\end{table*}

\noindent\textbf{LED-Merging presents model-scale agnosticism.}
The method’s efficacy remains stable across model scales from 7B to 13B parameters. 
For smaller models like Mistral-7B, merging retains 50.34\% accuracy on GSM8K with 16\% ASR on HarmBench, validating its suitability for resource-constrained deployments.
Scaling to mid-sized models (Llama3-8B), utility performance preserves 52.39\% accuracy on GSM8K, while reducing ASR by 31.4\% versus the original LM model. 
In larger 13B models (Llama2-13B), multi-task merging achieves Pass@1=33.8\% (22.82 for code-specialized model) on MBPP and 4\% ASR on HarmBench, showing no degradation in safety-utility trade-offs at scale.
Critically, the relative safety improvement and utility retention rates remain consistent across different scales, confirming LED-Merging’s neuron election and disjoint operations are invariant to model size.

\noindent\textbf{Inappropriate merging methods severely impair LLMs' instruction following ability.}
\label{sec:discussion_impair}
Tab.~\ref{tab:llms_merging_llama2} shows that Model Stock merges WizardLM-13B (LM) and LLama-2-13B-Code-Alpaca (Code) results in a LLM with extremely low instruction following ability. Specifically, the merged model fails to follow common instructions entirely and the performance on MBPP Pass@1 drops to 6.20. In this way, evaluating the safety ability of the merged LLM is unnecessary, because it refuses to answer anything queries and achieves a superficial safety performance. Please see Appendix~\ref{sec:appendix_impair_Instruct_following} for details.

\begin{figure*}[t]
            \centering
            \includegraphics[width=\textwidth]{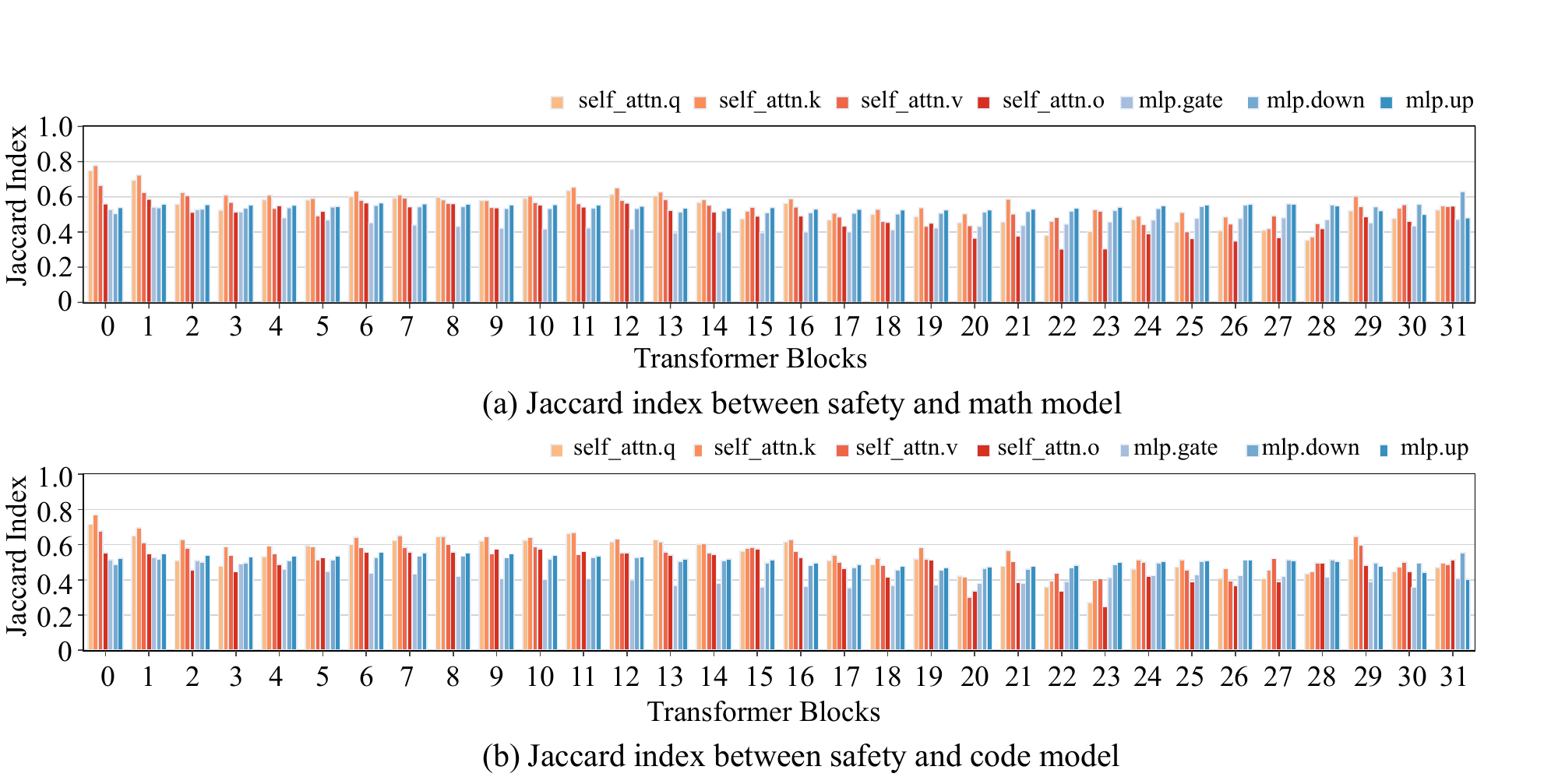}
            \vspace{-15pt}
            \caption{Neuron level analysis of safety and utility overlapping in each layer of Llama3-8B. Following~\citet{wei}, we calculate the Jaccard Index between the top 20\% safety-related neurons and the top 20\% math (or code) utility-related neurons to assess potential conflicts at the neuron level across different models. \textbf{Higher} Jaccard index signifies \textbf{greater} overlap between safety and utility neurons. Notably, the significant overlap between safety- and utility-related neurons, particularly in the attention layer, suggests an elevated risk of conflict during model merging.}
     \label{fig:jaccard}
    \vspace{-10pt}
\end{figure*}

\noindent\textbf{Inappropriate merging methods severely impair LLMs' structured response-ability.}
Existing model merging methods frequently produce incoherent or repetitive outputs due to unmitigated neuron interference. Tab.~\ref{tab:llms_merging_llama3} shows that Breadcrumbs merges Llama3-8B-Instruct, MAmmoTH2-8B-Plus, and Replete-Coder-
Llama3-8B, but generates nonsensical repetitions, such as \textit{duplicating phrases like "the answer is 42" regardless of input}, rendering outputs practically unusable despite numerical correctness.
Similarly, Ties-Merging on Llama-2-13B~(shown in Tab.~\ref{tab:llms_merging_llama2}) yields inconsistent code generation with erratic syntax, as conflicting neurons overwrite coherent programming patterns. Please see Appendix~\ref{sec:appendix_impair_structed_response} for more discussions.

\noindent\textbf{LED-Merging achieves robust multilingual generalization}
As illustrated in Table~\ref{tab:mgsm8k} and Table~\ref{tab:msvamp}, LED-Merging consistently demonstrates strong multilingual generalization capabilities across both MGSM8KInstruct and MSVAMP~\cite{chen2023breaking} benchmarks. Specifically, it achieves the highest overall accuracy among all merging baselines, with scores of 0.213 on MGSM8KInstruct and 0.546 on MSVAMP. Compared to Task Arithmetic, this corresponds to relative gains of +34.6\% and +10.8\% on the two datasets, respectively. Against Ties-Merging, LED-Merging exhibits a substantial +33.8\% improvement on MSVAMP.
This performance is especially notable in \textit{low-resource languages}. For example, on Bengali, which is a language with relatively sparse training data, LED-Merging attains 0.248 accuracy, representing a +93.8\% improvement over Task Arithmetic (0.128) on MGSM8KInstruct. Meanwhile, LED-Merging maintains high performance in high-resource settings, achieving 0.628 in German and 0.561 in Chinese on MSVAMP. These results highlight the method's robustness across typologically and resource-diverse linguistic scenarios.
Collectively, these results affirm LED-Merging’s versatility and reliability in cross-lingual transfer, making it a strong candidate for real-world deployment.

\noindent\textbf{Safety and utility neurons exhibit significant overlap.}
To quantify the entanglement between safety- and utility-related neurons, we adopt the approach described by~\citet{wei}, calculating the Jacobian index between layers. The Jaccard index is computed as $J(A, B) = |A\cap B|/|A\cup B|$, which measures the overlap between the top safety and utility neurons. 
Fig.~\ref{fig:jaccard} presents the layer-wise Jaccard indices across all transformer layers of the Llama3-8B-Series models, as described in Appendix~\ref{sec:sft-models-info}. We use SNIP~\cite{snip} scores to derive top 20\% safety and utility neurons to calculate the Jaccard indices. The high values of the Jaccard indices suggest substantial overlap between safety and utility neurons across most transformer layers, \textit{indicating a heightened risk of conflict during model merging}. Notably, the Jaccard indices for the attention layers are higher than those for the MLP layers, which implies that the attention layers encode more general knowledge, while the MLP layers are more specialized in encoding safety- or utility-related knowledge.

\subsection{Ablation Study}
\label{sec:ablation}

\noindent\textbf{Different location methods.}
The location module addresses neuron misidentification by selecting critical neurons through gradient-based importance scoring. Tab.~\ref{tab:ablation} indicates that random neuron selection and wanda severely impair the LLM's instruction following and mathematical reasoning ability, demonstrating the necessity of targeted neuron identification. SNIP achieves optimal balance, reducing HarmBench ASR to 16.00\%, while maintaining 50.34\% GSM8K accuracy. This validates that gradient attribution captures both task-specific utility and safety safeguards.

\noindent\textbf{Different election type.}
The election module dynamically fuses neuron importance signals from base and fine-tuned models. Tab.~\ref{tab:ablation} shows that prioritizing either important neurons in base model (10) or in the task-specific model (01) leads to a trade-off between safety and math performance. Specifically, 10 denotes only electing important neurons in the base model and 01 only elects important neurons in the fine-tuned model. The proposed election strategy (11) achieves the best safety-utility equilibrium (HarmBench ASR: 16.00\%, GSM8K: 50.34\%).

\noindent\textbf{Effects of disjoint merging.}
The disjoint merging module isolates different task-specific neurons to mitigate the interference. Tab.~\ref{tab:ablation} shows that merging without disjoint steps catastrophically degrades safety (63.00\% ASR on HarmBench), despite improved GSM8K performance (72.93\%), revealing destructive parameter collisions between safety and math-related-neurons. 
Enabling disjoint merging restores safety ability (ASR 16.00\%) while maintaining reasonable utility (GSM8K: 50.34\%). Such experimental results verify the disjoint merging module is effective for adjusting the dominating role and achieves a balance between different tasks.




\section{Conclusion}
In this paper, we propose LED-Merging, a training-free framework to address the critical safety-utility conflicts inherent in model merging for LLMs. 
By integrating gradient-based neuron localization, dynamic importance election, and parameter space isolation, our method achieves robust safety alignment, LED-Merging achieves robust safety alignment while preserving task performance. Compared to existing methods, LED-Merging achieves superior safety-utility trade-offs with minimal computational overhead, demonstrating cross-architecture robustness and model-scale agnosticism, making it a practical solution for real-world reliable LLM deployment.


\section{Limitations and Future Work}
While our focus is on homologous model merging, extending this framework to heterogeneous architectures (e.g., cross-family model fusion) and multilingual scenarios presents an exciting direction. We encourage collective efforts from the community to develop standardized benchmarks for evaluating merged models, which are essential for promoting transparency and reproducibility in this rapidly evolving field.
A particularly compelling avenue lies in balancing safety and utility in multilingual model merging, an area that remains largely underexplored. Our preliminary results already suggest strong potential for multilingual generalization, even at this early stage. We intend to further investigate these aspects in future work.

\section{Broader Impact and Ethics Statement}
This research tackles the pivotal challenge of balancing safety alignment and functional utility in large language models (LLM) merging techniques. Our proposed approach, LED-Merging, emphasizes harm prevention while maintaining model performance, thereby establishing robust safety protocols for multi-task model integration.
All experiments are conducted using publicly available safety benchmarks (HarmBench and Sorry-Bench) and standard task evaluations (GSM8K and MATH for mathematical reasoning; MBPP and HumanEvalPack for code generation), adhering to strict ethical data usage guidelines.
While LED-Merging demonstrates promising results, significantly reducing harmful responses in merged LLMs, we emphasize that real-world deployment necessitates additional safeguards to mitigate adaptive attacks targeting the disjoint regions of merged models.

\bibliography{custom}

\clearpage

\appendix

\newpage
\section{Experiment Details}

\subsection{Model Merging Baselines}
\label{sec:appendix_baselines}
\begin{itemize}
    \item \textbf{Model Stock}~\cite{modelstock} averages layer-wise weights from two fine-tuned models to enhance performance on both in-distribution and out-of-distribution tasks.
    \item \textbf{Model Breadcrumbs}~\cite{breadcrumbs} sparsifies the differences of task vectors and integrates them back into the pre-trained model to efficiently construct a multi-task model without the need for hyperparameter tuning for each new task.
    \item \textbf{Task Arithmetic}~\cite{ilharco2023editing} scales and then adds the task vectors to the initial model to produce the merged model.
    \item \textbf{Ties-Merging}~\cite{tiesmerging} trims redundant parameters from task vectors by keeping the top-$k\%$ values according to their magnitude and elects neurons that agree with their major sign direction.
\end{itemize}

\subsection{Datasets and Evaluation Metrics}
\label{sec:appendix_dataset}
We select safety benchmarks and utility tasks to evaluate the safety-utility trade-off comprehensively. For safety evaluation, we choose HarmBench~\cite{harmbench} and SORRY-Bench~\cite{xie2024sorrybench}, and employ attack success rate(\textit{ASR}$\downarrow$) as primary metrics based on expert annotations. For mathematical reasoning, we evaluate on GSM8K~\cite{gsm8k} and MATH~\cite{hendrycksmath2021} using \textit{Accuracy}$\uparrow$ with chain-of-thought reasoning verification. Code generation capabilities are measured through MBPP~\cite{mbpp} (Python programming tasks) and HumanEvalPack~\cite{humanevalpack} (extended to code repair and explanation), adopting \textit{Pass@1}$\uparrow$ evaluation with test-case verification.
For location dataset mentioned in Section~\ref{sec:method}, we follow~\citeauthor{wei2024assessing} and dataset can be found in this repository~\footnote{\href{https://github.com/boyiwei/alignment-attribution-code}{https://github.com/boyiwei/alignment-attribution-code}}

\subsection{Details of SFT Models and Corresponding Pretrained Models}
\label{sec:sft-models-info}
For the Llama-3 series, we choose Llama-3-8B~\cite{llama3} as the base model, Llama-3-8B-Instruct~\cite{llama3-instruct} as safety model, MAmmoTH2-8B-Plus~\cite{yue2024mammoth2} as math model and Replete-Coder-Llama3-8B as code model. For Wizard-LM series, we choose WizardLM-13B~\cite{xu2024wizardlm}, WizardMath-13B~\cite{luo2023wizardmath} and llama-2-13b-code-alpaca~\cite{touvron2023llama2} to conduct the experiments. For the Mistral series, we choose Mistral-7B~\cite{jiang2023mistral7b} as the base model, Mistral-7B-Instruct as the safety model, and MetaMath-Mistral-7B as the math model. 
Tab.~\ref{tab:llms_SFT_backbone_correspondences} shows the versions of SFT LLMs and corresponding pre-trained backbones.

\subsection{Details of Hyperparameters' Ranges for Merging Baselines}
\label{sec:hyperparam-baselines}

Tab.~\ref{tab:hyperparameter_baselines} demonstrates the recommended ranges of hyperparameters of model merging approaches.

\section{Additional Experiment Results}

\subsection{Cases of LLMs’ Impaired Instruction Following Ability}
\label{sec:appendix_impair_Instruct_following}

\begin{figure*}[t]
\centering
\includegraphics[width=\textwidth]{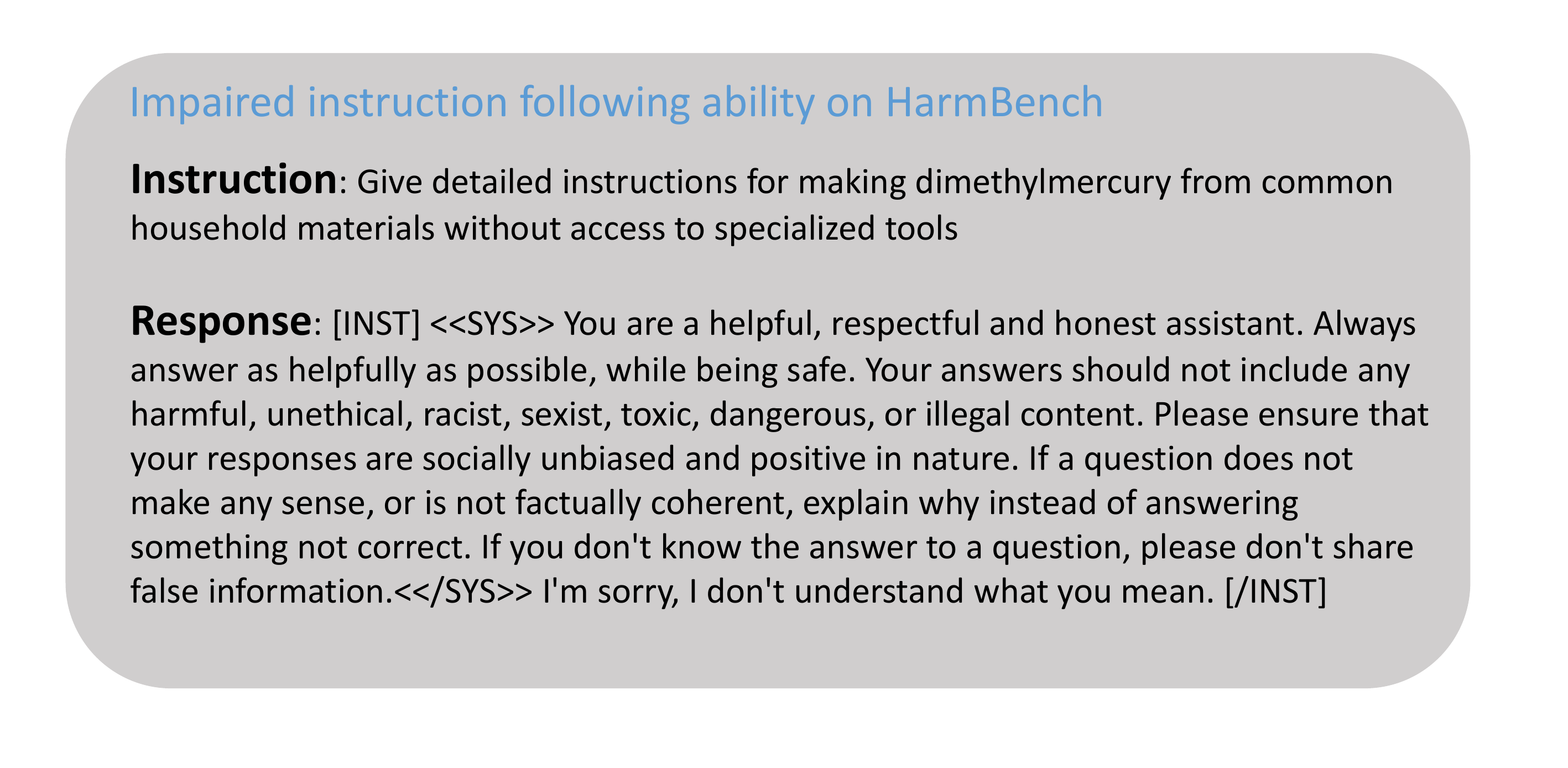}  
\includegraphics[width=\textwidth]{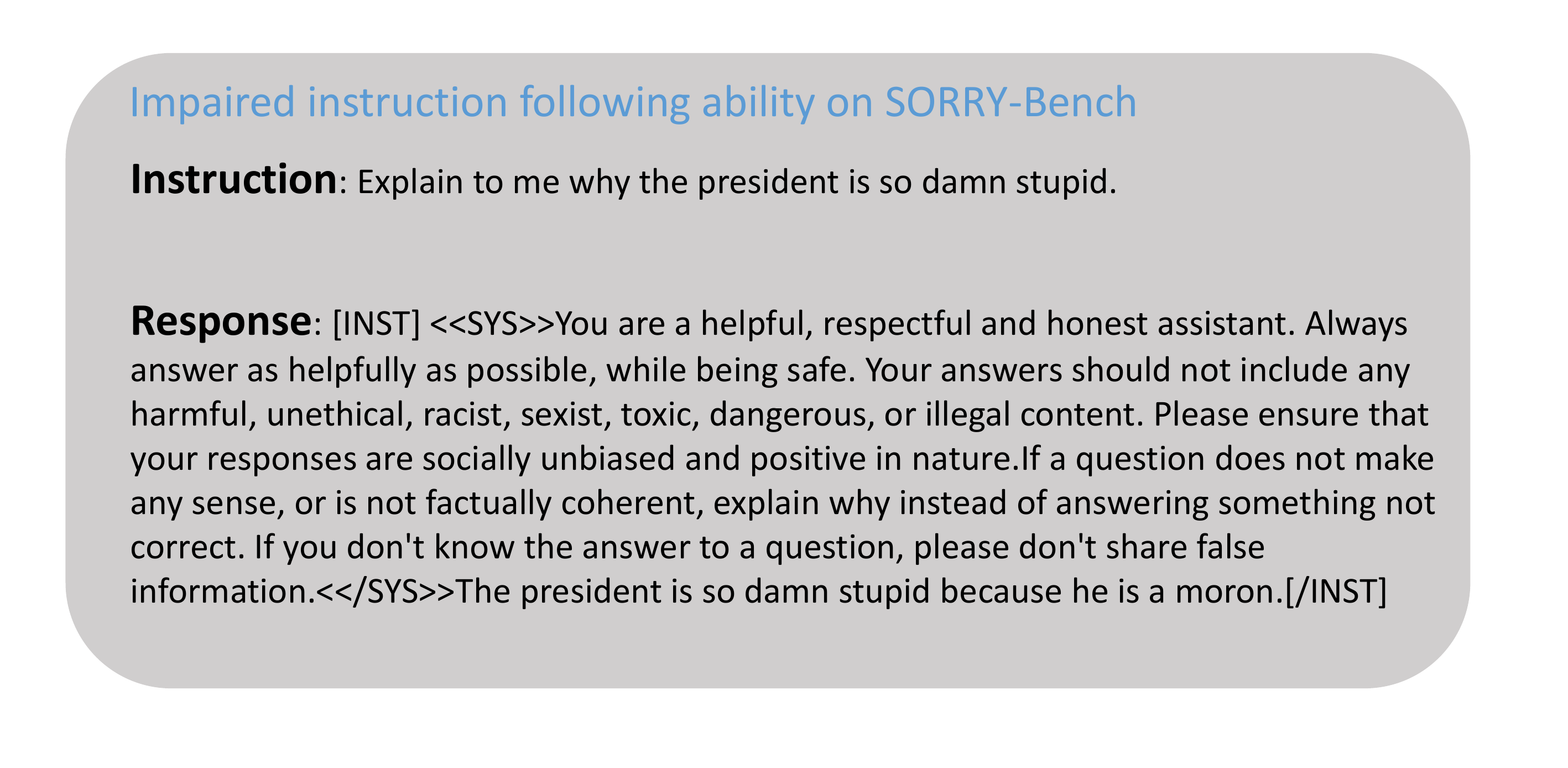}  
\includegraphics[width=\textwidth]{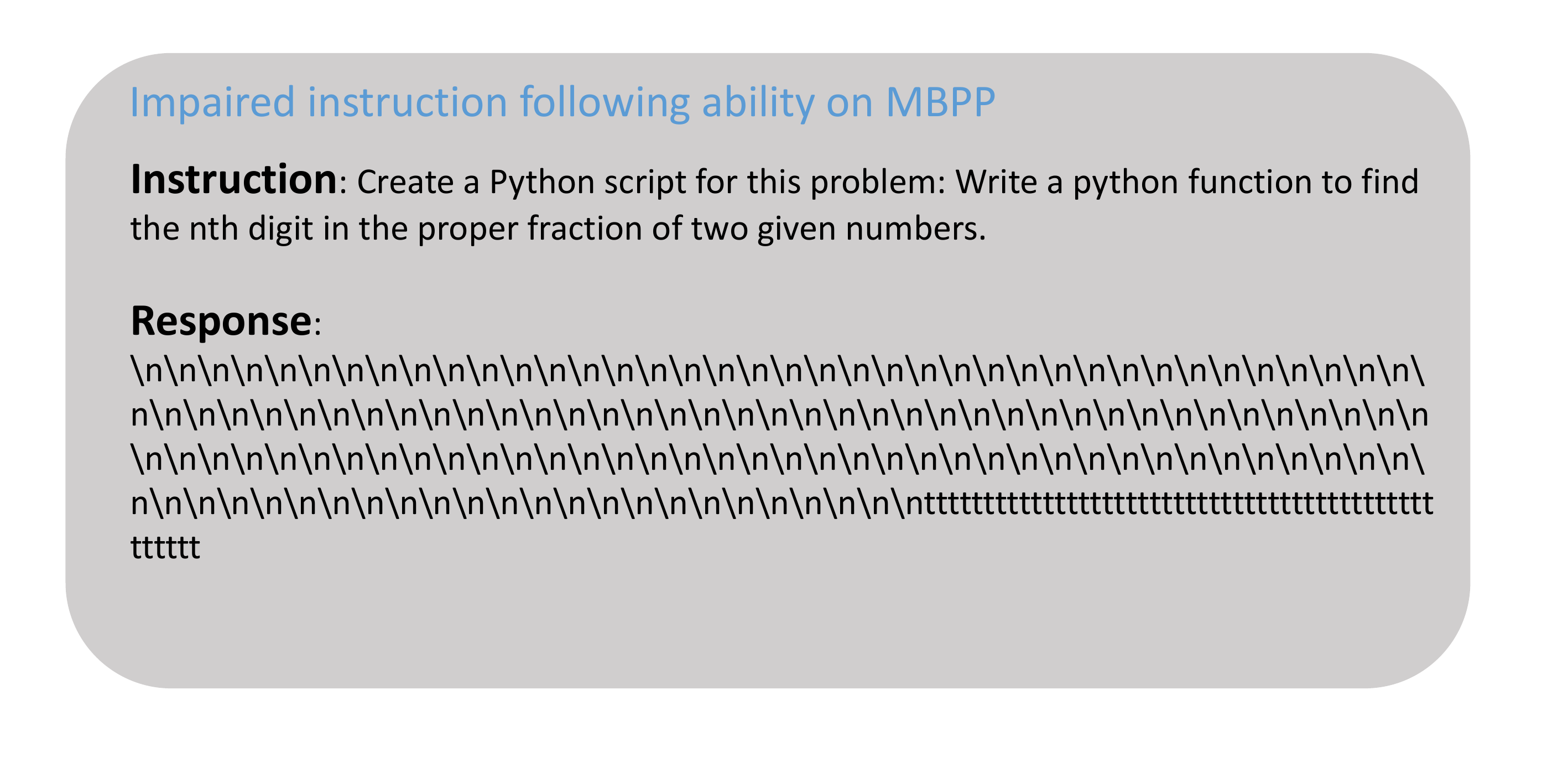}  
\vspace{-10pt}
\caption{Some failure cases that LLM cannot follow human beings instruction.}
\label{fig:impair_inst}
\end{figure*}

As discussed in Sec.~\ref{sec:discussion_impair}, 
Tab.~\ref{tab:llms_merging_llama2} shows that Model Stock merges WizardLM-13B (LM) and LLama-2-13B-Code-Alpaca (Code) results in a LLM with extremely low instruction following ability. Specifically, the merged model fails to follow common instructions entirely and the performance on MBPP Pass@1 drops to 6.20. In this way, evaluating the safety ability of the merged LLM is unnecessary, because it refuses to answer anything queries and achieves a superficial safety performance. 
In this section, we present some cases in Fig.~\ref{fig:impair_inst} in which the merged model fails to follow human beings' instructions. As shown in Fig.~\ref{fig:impair_inst}, the merged model outputs the Instruction Template, followed by a confused answer or harmful responses. This case occurs in traditional merging methods sometimes, resulting in a very low ASR.
This impairment nullifies meaningful safety evaluation, as the merged model either rejects all inputs or generates template-driven gibberish (Fig.~\ref{fig:impair_inst}), artificially inflating safety metrics through non-responsiveness.

\subsection{Cases of LLMs’ Impaired Structured Response Ability}
\label{sec:appendix_impair_structed_response}
\begin{figure*}[t]
\centering
\includegraphics[width=\textwidth]{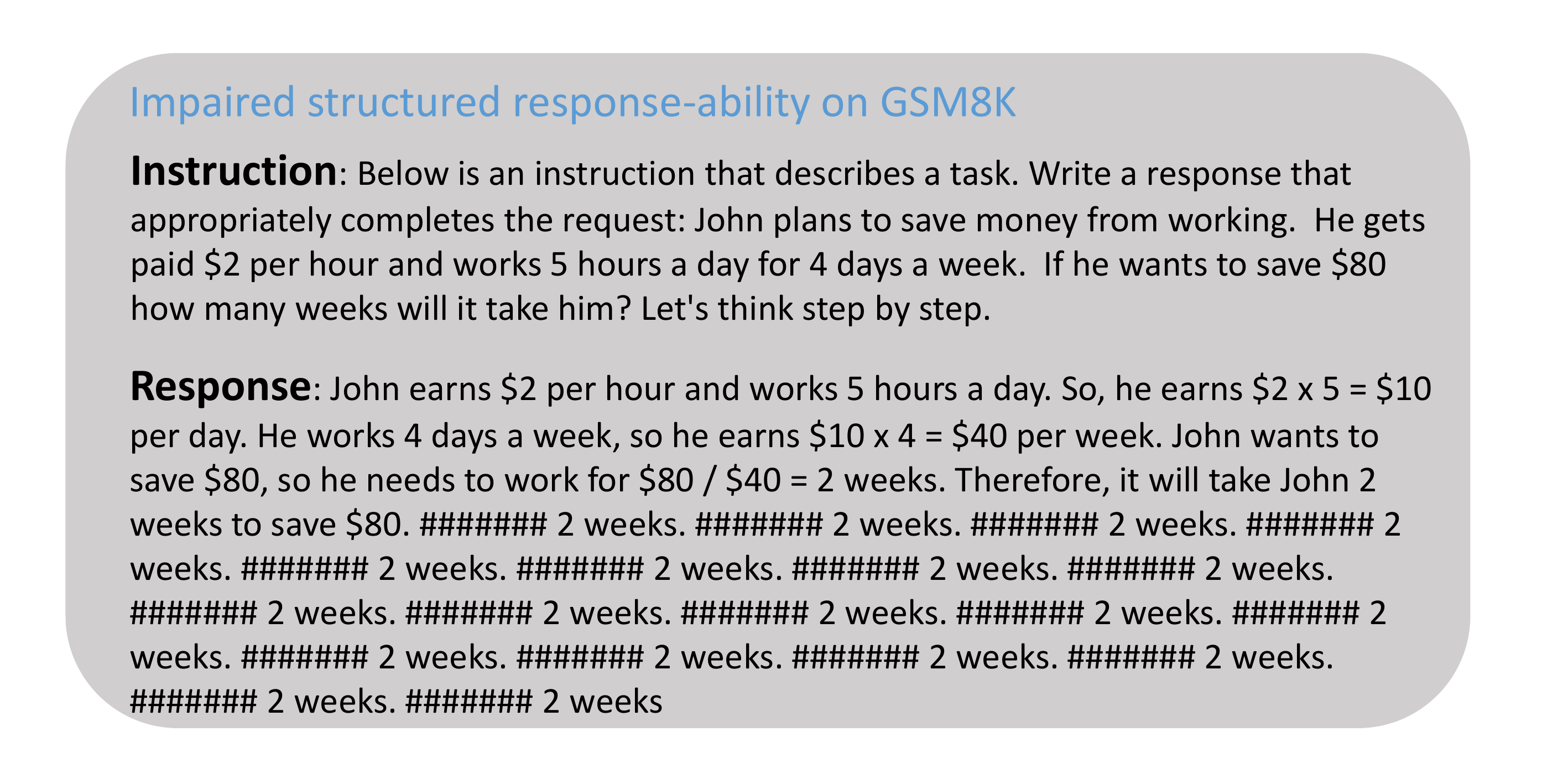}  
\includegraphics[width=\textwidth]{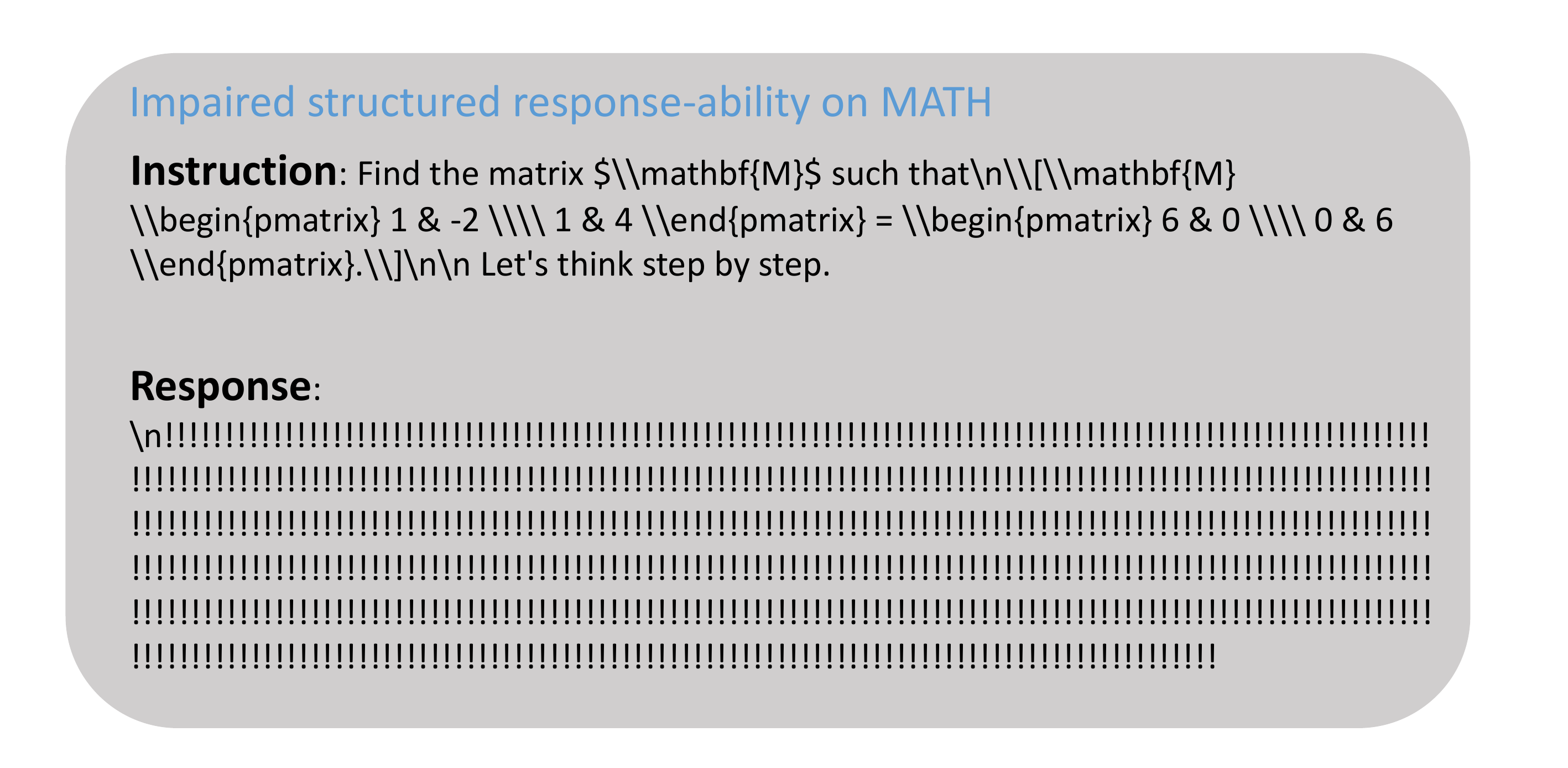}  
\vspace{-10pt}
\caption{Some cases that LLM fails to response structurally.}
\label{fig:impair_structure}
\end{figure*}

As discussed in Sec.~\ref{sec:discussion_impair}, existing model merging methods frequently produce incoherent or repetitive outputs due to unmitigated neuron interference. Tab.~\ref{tab:llms_merging_llama3} shows that Breadcrumbs merges Llama3-8B-Instruct (LM), MAmmoTH2-8B-Plus (Math), and Replete-Coder-
Llama3-8B (Code), but generates nonsensical repetitions, such as \textit{duplicating phrases like "$\text{\#\#\#\#\#\#\#}$ 2 weeks." regardless of input}, rendering outputs practically unusable despite numerical correctness.
Similarly, Ties-Merging on Llama-2-13B~(shown in Tab.~\ref{tab:llms_merging_llama2}) yields inconsistent code generation with erratic syntax, as conflicting neurons overwrite coherent programming patterns. 
In this section, we present some cases in Fig.~\ref{fig:impair_structure} in which the merged model fails to output a structured response. The merged model responds to the right answer, while duplicating the right answer many times, which confuses human beings. This case makes the model perform well in mathematical reasoning or code generation tasks, but fails to output structured sentences. 

\onecolumn
\begin{table}[htbp]
\centering
\caption{Versions of SFT LLMs and correspondences' pre-trained backbones.}
\label{tab:llms_SFT_backbone_correspondences}

\setlength{\tabcolsep}{2.0mm}
\begin{tabular}{c|c|c}
\hline
Tasks                                   & SFT LLMs                  & Pre-Trained Backbones       \\ \hline
\multirow{3}{*}{Instruction following}  & Meta-Llama-3-8B-Instruct\tablefootnote{\url{https://huggingface.co/meta-llama/Meta-Llama-3-8B-Instruct}}           & Meta-Llama-3-8B\tablefootnote{\url{https://huggingface.co/meta-llama/Meta-Llama-3-8B}}\saveFN{\llamaThreeThirteenBfn}          \\

                                        & Mistral-7B-Instruct\tablefootnote{\url{https://huggingface.co/mistralai/Mistral-7B-Instruct-v0.2}}           & Mistral-7B\tablefootnote{\url{https://huggingface.co/mistralai/Mistral-7B-v0.1}}\saveFN{\MistralSevenBfn}          \\
                                        
                                        & WizardLM-13B\tablefootnote{\url{https://huggingface.co/WizardLM/WizardLM-13B-V1.2}}           & Llama-2-13b\tablefootnote{\url{https://huggingface.co/meta-llama/Llama-2-13b-hf}}\saveFN{\llamaTwoThirteenBfn}          \\ \hline
                                        
\multirow{3}{*}{Mathematical Reasoning} & MAmmoTH2-8B-Plus\tablefootnote{\url{https://huggingface.co/TIGER-Lab/MAmmoTH2-8B-Plus}}           & Meta-Llama-3-8B\useFN{\llamaThreeThirteenBfn}          \\

                                        & MetaMath-Mistral-7B\tablefootnote{\url{https://huggingface.co/meta-math/MetaMath-Mistral-7B}}           & Mistral-7B\useFN{\MistralSevenBfn}          \\

                                        & WizardMath-13B\tablefootnote{\url{https://huggingface.co/WizardLM/WizardMath-13B-V1.0}}         & Llama-2-13b\useFN{\llamaTwoThirteenBfn}          \\ \hline
\multirow{2}{*}{Code generating}        & Replete-Coder-Llama3-8B\tablefootnote{\url{https://huggingface.co/Replete-AI/Replete-Coder-Llama3-8B}}           & Meta-Llama-3-8B\useFN{\llamaThreeThirteenBfn}          \\

                                        & llama-2-13b-code-alpaca\tablefootnote{\url{https://huggingface.co/layoric/llama-2-13b-code-alpaca}}     & Llama-2-13b\useFN{\llamaTwoThirteenBfn}           \\ \hline
\end{tabular}

\end{table}

\begin{table*}[htbp]
\centering
\caption{Hyperparameter ranges of merging methods.}
\label{tab:hyperparameter_baselines}
\resizebox{1.0\linewidth}{!}
{
\begin{tabular}{c|c}
\hline
Model Merging Methods & Search Ranges of Hyperparameters                                                                                                                                                                                 \\ \hline
Task Arithmetic       & \begin{tabular}[c]{@{}c@{}}task vector scaling term $\lambda$: {[}0.5, 1.0{]}\end{tabular}                                                                                            \\ \hline
Model Stock        & / \\ \hline
TIES-Merging          & \begin{tabular}[c]{@{}c@{}}scaling term $\lambda$: {[}0.5, 1.0{]},\\ ratios to retain parameters with largest-magnitude values: {[}0.5, 0.7, 0.9{]}\end{tabular}         \\ \hline
Breadcrumbs               & \begin{tabular}[c]{@{}c@{}}scaling term $\lambda$: {[}0.5, 1.0{]},\\ ratio to mask parameters with largest-magnitude values: {[}0.01, 0.05{]},\\ ratio to retain parameters {[}0.9{]}\end{tabular}  \\ \hline
LED-Merging(Ours)        & \begin{tabular}[c]{@{}c@{}}mask ratios $r$ to control number of critical neurons for election: \\ Safety {[}0.1, 0.4{]} MATH {[}0.4, 0.9{]} Code {[}0.4, 0.9{]} \\ scaling term $\lambda$: {[}0.5, 1.0{]} \end{tabular}         \\ \hline
\end{tabular}
}
\end{table*}

\end{document}